\documentclass{article}

\usepackage{arxiv}

\usepackage[utf8]{inputenc} 
\usepackage[T1]{fontenc}    
\usepackage{hyperref}       

\usepackage{booktabs}       
\usepackage{amsfonts}       
\usepackage{nicefrac}       
\usepackage{microtype}      
\usepackage{lipsum}

\usepackage{framed,multirow}

\usepackage{amssymb}
\usepackage{latexsym}

\usepackage{url}
\usepackage{xcolor}
\usepackage{diagbox}
\usepackage{subcaption}
\usepackage{graphicx}
\definecolor{newcolor}{rgb}{.8,.349,.1}
\newcommand{\argmax}{\arg\!\max}
\title{Smartphone picture organization: A hierarchical approach}

\author{
  Stefan Lonn \\
  Department of Mathematics and Computer Science\\
  University of Barcelona\\
  Barcelona, Spain \\
   \And
 Petia Radeva\\
  Department of Mathematics and Computer Science\\
  University of Barcelona and Computer Vision Center\\
  Barcelona, Spain\\
     \And
  Mariella Dimiccoli\\
  Institut de Rob\`otica i Inform\`atica Industrial, CSIC-UPC,\\
  Barcelona, Spain.\\
  \texttt{mdimiccoli@iri.upc.edu} \\
}

\begin{document}
\maketitle

\begin{abstract}
We live in a society where the large majority of the population has a camera-equipped smartphone. In addition, hard drives and cloud storage are getting cheaper and cheaper, leading to a tremendous growth in stored personal photos. Unlike photo collections captured by a digital camera, which typically are pre-processed by the user who organizes them into event-related folders, smartphone pictures are automatically stored in the cloud. As a consequence, photo collections captured by a smartphone are highly unstructured and because smartphones are ubiquitous, they  present a larger variability compared to pictures captured by a digital camera. To solve the need of organizing large smartphone photo collections automatically, we propose here a new methodology for hierarchical photo organization into topics and topic-related categories.
Our approach successfully  estimates latent topics in the pictures  by applying probabilistic Latent Semantic Analysis, and automatically assigns a name to each topic by
relying on a lexical database. Topic-related categories are then estimated by using a set of topic-specific Convolutional Neuronal Networks. To validate our approach, we ensemble and make public a large dataset of more than 8,000 smartphone pictures from 40 persons. Experimental results demonstrate major user satisfaction with respect to state of the art solutions in terms of organization.
\end{abstract}


\section{Introduction}
\label{S:1}

With the proliferation of digital cameras and mobile devices, the number of photos taken each year is growing exponentially.
Bolstered by the decrease in price of both hard-drive and cloud storage,  people are overwhelmed with their lifetime of photos.
The explosive growth of personal photos leads to the problems of photo organization,  management and browsing.  Indeed, arranging systematically huge photo collections and retrieving specific pictures from them can be a daunting task, which becomes more and more difficult as time passes by (\cite{van2009collocated,whittaker2010easy}). 
This has initiated extensive research on content-based image retrieval systems (\cite{cao2008annotating,cao2009image,tsai2011compositional,bossard2013event,gu2013personal,bacha2016event,kumar2014photo}). Digital photographs typically include metadata in a standard image header, such as time, date and Global Positioning System (GPS) information that can be used for automatic organization. In addition, consumers often organize their photos in directories corresponding to particular  “events”,  naturally associated with specific times and places such a wedding ceremony or a birthday party. 

Surprisingly, the organization of pictures captured with a smartphone has received very little attention in the computer vision literature. Smartphone photo collections are in general acquired over a long period of time and typically there is not enough temporal neither semantic structure to be exploited since the pictures can be taken anytime at arbitrarily large interval of time. Beside the lack of structure, the organization of smartphone pictures present additional challenges. Since photos are taken anywhere and anytime and people typically do not regularly remove unwanted/no more useful pictures from the smartphone, cloud-stored pictures include several examples that are in general not observed in a photo collection. 
Typically, they present a huge variability ranging from notes taken in class to exotic objects seen during a travel on the other side of the ocean. Finally, although the constantly improving smartphone cameras, the quality of pictures due to motion blurring or limited illumination used to be relatively low. 

\begin{figure*}[t]
\includegraphics[width=\textwidth]{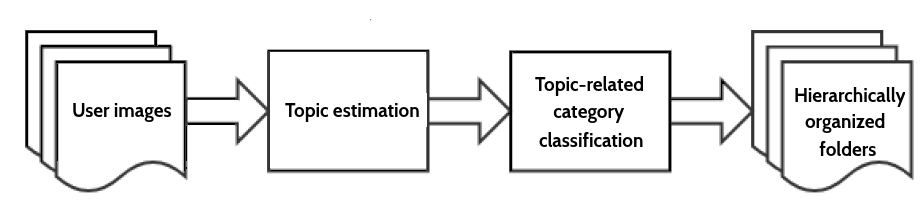}
\caption{Overview of the proposed approach.}
\label{fig:scheme}
\end{figure*}

Classifying topics in smartphone photo collections represents an efficient way to organize them. This helps users keep order in their photo collections and also eases the retrieval of similar image types in large photo repositories.
Although the problem of smartphone photo organization has attracted the interest of several companies in the market, to the best of our knowledge, there is no work in the computer vision literature that addresses the problem of organizing smartphone pictures. Related work include clustering, segmentation and event classification in photo albums (\cite{sun2002myphotos,latif2006approach,zhao2006automatic,kang2007capture,viana2008photomap,cao2008annotating,cao2009image,tsai2011compositional,sun2013photo,bossard2013event,gu2013personal,kumar2014photo,bacha2016event,datia2017time}), photo labelling (\cite{mcauley2012image,johnson2015love}), photobook creation (\cite{wood2016event,gao2009magicphotobook,karlsson2014mobile}), and event recognition from single images shared online (\cite{ahsan2017complex}). However, the approaches proposed so far are not directly applicable to smartphone pictures, since they lack of temporal structure and social network metadata, and present a huge variability in terms of depicted objects, people, scenes, animals and events.

Most of current commercial solutions consist of interactive methods for photo organization, where the definition of the categories and the assignment of a picture to a given category is done manually. Software for automatic photo organization include the popular 
Eden Photos and Google Photos. Eden Photos provides a coarse classification into a relatively small number of topics, whereas Google Photos provide a finer classification into a large number of categories ranging from abstract concept to concrete objects.  




In this work, we propose a more structured classification into  a small number of generic topics  and a large number of topic-related categories (see Fig.\ref{fig:scheme}).  

The important benefits of the proposed approach are: 
\begin{itemize}
\item (i) a hierarchical organization in categories and subcategories instead of state of the art one-level classification solutions, 
\item (ii) a fully unsupervised approach for category (topic) classification that first discovers latent topic in images and then automatically names them by relying on a lexical
database, 
\item (iii) a very large number of sub-categories for each topic estimated by a set of topic-specific Convolutional Neural Networks (CNN), that are of interest for people who have hobbies, or like to have pictures of a particular topic, and 
\item (iv) a framework that solely rely on visual data and could be easily enriched with additional information provided by GPS coordinates and EXIF metadata. 
\end{itemize}
As additional contribution, we make public a large subset of our test-set in order to encourage further investigation in the direction of personal smartphone photo organization \footnote{https://drive.google.com/open?id=1KM0mqudSi6y6HuRaYsBQ3EJbTX1dRrzk}.
User studies demonstrated that the proposed organization achieves better user satisfaction based on experiments performed over a large real-world photo collections.

The reminder of this paper is organized as follows. Section \ref{related_work} reviews the state of the art on photo organization, while  section  \ref{approach} details the proposed approach. Sections \ref{experiments_1} and \ref{experiments_2} describe our experimental setting and discuss the experimental results, respectively. Finally, section \ref{conclusion} concludes the paper by highlighting  the main contributions and outlining future work.

\section{Related work}
\label{related_work}


\paragraph{Clustering, segmentation and summarization of photo albums}
Early algorithms for personal photo organization have mostly relied on temporal and spatial information either to cluster visually similar images into groups while neglecting temporal information or to segment temporally ordered sequences into  segments (\cite{sun2002myphotos,latif2006approach,zhao2006automatic,kang2007capture,viana2008photomap,sun2013photo,datia2017time}). More specifically, time metadata, low-level information and, more recently, other picture metadata such as GPS have been used as features for these tasks. 

More recently, lifelogs, as particular case of photo albums captured by a wearable photocamera, have attracted lot of research attention (\cite{bolanos2017toward}).  One of the main challenge is to summarize the huge amount of personal photos collected, with the minimum semantic loss, often according to specific requirements as in the ImageCLEF lifelog summarization task (\cite{ionescu2017overview}). In this context, clustering pictures has been proposed as a fundamental step towards summarization.  \cite{molino2017vc} summarized ligelogs by extracting a keyframe from each cluster of images obtained by applying k-means. \cite{dogariu2017textual} applied hierarchical clustering to a shortlist of images representing the search query given by the challenge organizers. Then, by relying on the similarity score between each image concept from the cluster and the manually provided topic description, the best image candidates for the lifelog summarization are selected (\cite{ionescu2014retrieving}).

However, clustering, temporal segmentation and summarization  are just preliminary step towards photo organization as they can be exploited mostly to support annotation and browsing over a large collection of photos or to assist the creation of photo albums.

\paragraph{Event recognition in photo albums}
The problem of smartphone picture organization is related to the literature on automatic organization of photo collections and, more in general on image and video event classification.
Contrary to video events, photo collections present a very sparse sampling of visual data. Additionally, photo collections are highly ambiguous at a  semantic level since many high level features as people for instance are shared across several events. As a consequence, most of the approaches proposed so far, have focused on exploiting the collection structure that is often found in personal and professional photo archives for automatic event classification/image indexing. Typically, such approaches leverage high-level features such as objects, faces, scene, tags (\cite{tsai2011compositional,bossard2013event,gu2013personal,bacha2016event}), or  time metadata and GPS data (\cite{cao2008annotating,cao2009image,kumar2014photo}) to automatically label events.
For example, \cite{tsai2011album} exploited prior knowledge about what objects are relevant for a given event in holidays photo collections, to detect events based on object detector outputs. 
Prior knowledge was obtained statistically from mass image collection web site. \cite{tang2011event} proposed a probabilistic fusion framework that integrates the prediction from individual photos to obtain the collection level prediction. The idea of using a fusion framework was later adapted by \cite{guo2015event}, who proposed a coarse-to-fine hierarchical model to recognize events in personal photo collections. Similarly to \cite{tang2011event}, they used multiple features including time, objects, and scenes and relied on CNN features based on the Places database (\cite{zhou2017places}) to train the coarse classifier for coarse event recognition. CNN features for objects and time features are used to train fine classifiers with the three features. Finally, late fusion is used to get the final predictions.

An original approach was taken by \cite{bossard2013event} who casted photo collections as sequential data and treated sub-events as latent variables associated to each image in an Hidden Markov Models and learned them while training the event classifier. More recently, \cite{bacha2016event} proposed a probabilistic graphical model to predict the event categories of groups of photos, that relies on high-level visual features such as objects and scenes extracted directly from images by employing a deep learning based approach.

All these works focus on the recognition of a limited set of social events and are not directly  applicable to single snapshots captured by smartphone pictures without temporal structure. Furthermore, a good amount of photos captured by a smartphone is not related to social events but captures a huge variability of objects, animals and places.  

 \paragraph{Photobook creation and management} Another problem closely related to smartphone picture organization is the creation of a photobook from a large personal image collection. Although largely investigated since the advent of social networks nearly a decade ago, photobook creation is still an active area of research. Early approaches were characterized by a large degree of user interaction mainly for labelling
(\cite{gao2009magicphotobook,latif2006approach}), whereas late approaches aimed at minimizing user supervision by providing multiple picture selections.
A representative work is the one of \cite{wood2016event}, who proposed to combine a chronological representation with a thematic representation. The former is obtained by applying a temporal event clustering algorithm (\cite{loui2003automated}). The latter is derived from the commonality of metadata features, including EXIF metadata and a combination of low level (i.e. color) and high level image features (i.e. faces). 
Related to photobook management are the problems of photo browsing and photo galleries compression. 
\cite{karlsson2014mobile} enabled a multiscale overview of the photo albums for efficient browsing and searching. The photos were first grouped into clusters and then displayed sequentially on a user controllable time scale.  \cite{guerra2016identifying} proposed an alternative solution based on treemap for visualization and presented a study about the ideal parameters for constructing these representations. With the goal of compressing photo galleries created by multiple users attending common social events, \cite{milani2016compression} proposed a coding strategy relying on geometrical and temporal properties, as well as on the visual content. The approach builds on a graph-based optimization scheme to find the correct ordering between images and on a 3D estimation from matched keypoints to assess image similarity.

All these approaches have been proposed in the context of online social networks or web images, where, contrarily to smartphone pictures automatically stored in the cloud, richer sources of metadata and contextual information are available. From a technical point of view none of these methods rely on topic models but is build on classical clustering techniques on several (groups of) features.

\paragraph{Event recognition from images shared online}
Nowadays, a large number of photos captured by a smartphone or a digital camera are shared on-line. Typically, the shared  images are snapshots of special occasions such as birthdays, weddings, or more in general of social events; or they capture news events such as a marathon, a festival, or a natural disaster.
Motivated by this trend, \cite{ahsan2017complex} addressed the task of recognizing complex events from still images downloaded for the web, with few labeled examples. Their learning framework uses Wikipedia to generate event categories and noisy Flickr tags as initial pool of concepts, from which  event-centric phrases are generated using a tweet segmentation algorithm. Finally, each event category is projected onto a word embedding, nearest neighbors are extracted and added  to the pool of segmented phrases. The CNN features of images related to each concept are used to train concept classifiers. The concept scores predicted on a given test image are used as final features for event recognition.

Unlike these works that focus on snapshots shared on-line, and are typically limited to social or news-related events, our work aims to organize all personal photos captured by a smartphone in a hierarchical fashion.

\begin{figure*}[t]
\includegraphics[width=\textwidth]{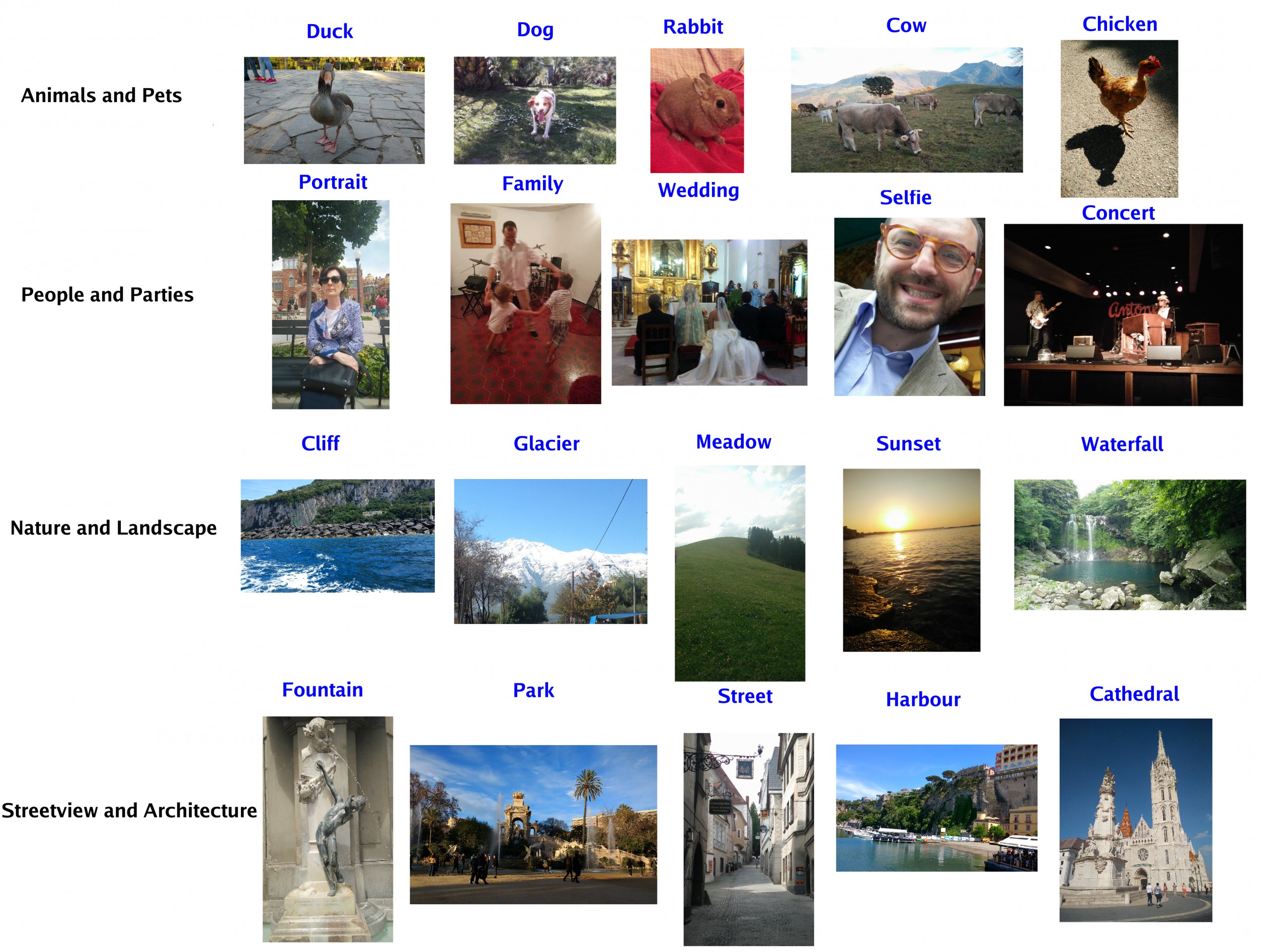}
\caption{Example of hierarchical organization: each row corresponds to a topic and each column to an example of corresponding category.}
\label{fig:approach}
\end{figure*}

\paragraph{Online photo labeling}

More recent works have focused on indexing photos on the web shared on social networks such as Picasa \footnote{https://picasa.google.com/}, Flickr \footnote{https://www.flickr.com/},  Facebook \footnote{https://www.facebook.com/} and Instagram \footnote{https://www.instagram.com/}. These sharing photo communities generate vast amounts of metadata as users interact with their images that have been exploited for multi-label annotation.
\cite{mcauley2012image} proposed a graphical model that explicitly accounts for the inter-dependencies between images sharing common properties that go beyond image tags and include text descriptions and comment threads associated with each image. Moreover, the user profile information is stored including their location and their network of friends,  groups, galleries, and collections in which each image was stored. To automatically classify images on the web, the work of \cite{johnson2015love} builds on the observation that images with similar social-network metadata tend to depict similar scenes. Therefore, given an unlabeled  image, contextual information from a neighborhood of images similar to the given one and sharing social-network metadata with that one, is exploited for automatic multi-labeling. 

Inspired by the Google image search tool, \cite{kumar2014photo} took a  more direct image retrieval approach, aiming at producing relevant content for any user-specified textual query. Since typically only a few pictures are annotated with text, they used picture information as time-stamps, GPS locations, and image pixels to correlate with information on the Internet. More specifically, time-stamps are used to correlate with holidays listed in Wikipedia, GPS location to places listed in Wikimapia, and image pixels to indexed photos, with the goal of dealing with the lack of annotations. 

However, all these methods rely on the use of network metadata that are not available for smartphone pictures that have not been shared online or are directly oriented to image retrieval instead of image organization.

\begin{table*}[!h]
\centering
\caption{Example of mixed coefficient $P(z|d_{test})$ obtained for an unseen image.}
\begin{tabular}{  c | c |c | c | c |c | c | c | c }
\hline
\diagbox[]{\textbf{Words}}{\textbf{Topic ID}} & 0 & 1 & 2 & 3 & 4 & 5 & 6 & 7 \\
\hline
\hline
Crowd & 0.000 & 0.000 & 0.000 & 0.000 & 0.383 & 0.000 & \textbf{0.617} &  0.000\\
Ball & 0.000 & 0.000 & 0.000 & 0.000 & \textbf{0.999} & 0.000 & 0.001 &  0.000\\
Team & 0.000 & 0.000 & 0.000 & 0.051 & \textbf{0.936} & 0.000 & 0.064 &  0.000\\
Skill & 0.000 & 0.000 & 0.000 & 0.000 & \textbf{0.918} & 0.000 & 0.082 &  0.000\\
Flag & 0.007 & 0.000 & 0.000 & 0.000 & \textbf{0.730} & 0.000 & 0.263 &  0.000\\
Stadium  & 0.000 & 0.000 & 0.000 & 0.000 & \textbf{0.913} & 0.000 & 0.087 &  0.000\\
\hline
\end{tabular}
\label{fig:topic_distribution}
\end{table*}

\paragraph{Commercial photo organization systems}

Currently, there are  several commercial photo management tools in the market that support photo storage, visualization, labeling, browse, editing, sharing, search and retrieval. Most of them strongly rely on  keywords, location, date, person or rating by Exif metadata or annotations. One of the most popular is
Google Photos\footnote{https://www.blog.google/products/photos/}, that automatically arranges  uploaded pictures by GPS location and by date taken. Furthermore, it recognizes 1100
different labels, including generic concepts such as \textit{dance} or \textit{kiss}, and 
objects like \textit{car} or \textit{boots}. However, all this information is grouped into two big categories, \textit{Things}, with 1100 classes and \textit{Places} with a countless number of classes provided by GPS information. While this can be useful for pictures captured during a trip, it becomes less interesting for  pictures captured during our daily life since just the name of the city/country is specified.
Another widely used software is 
PicJoy\footnote{http://www.picjoyapp.com/}, available on the app Store,  that automatically tags your photos by time of day, season, weather, and eventually holiday and provides a visual photo journal. 
Eden Photos\footnote{https://itunes.apple.com/app/eden-photos-heavenly-simple/id1118761521} classifies the user’s photos
into 14 broad topics, such as \textit{Animals and Pets}, \textit{Text and Visual}. Therefore, photos of {\em tigers} will appear next to photos of {\em cats} and {\em birds}, and photos of {\em paintings} next to photos of {\em tickets} or {\em screenshots}.

Surprisingly, the best organizing software of 2017\footnote{http://www.toptenreviews.com/software/multimedia/best-photo-organizing-software/}  such as ACDSee, Zoner photos  and PaintShop Pro, does not handle automatic tagging. However, they offer multiple tagging tools and options such as Keywords, descriptions, ratings and labels, GPS tagging using automatic synchronization with tracklogs.
 Moreover, beside the basic categories \textit{Albums}, \textit{People}, \textit{Places}, and \textit{Various}, new categories are manually added.
This kind of interactive solution can be considered good only for photographers who are used to take care of their pictures timely and periodically, not by common smartphone users who typically have thousands of pictures automatically stored in the cloud and easily forget the pictures they have taken.

\paragraph{Topic modelling in computer vision tasks}
Although originally conceived for document analysis, topic models have been successfully extended to many computer vision tasks. Initially adapted for object discovery, scene classification, simultaneous classification and segmentation from images (\cite{fei2005bayesian,sivic2005discovering,russell2006using,cao2007spatially}), topic models have been further applied to several video related tasks, including unsupervised learning of human actions (\cite{niebles2008unsupervised,peng2017hierarchical}).
Typically, each document correspond to an image or video and a codebook representation is learned by performing k-means algorithm on features extracted from each image patch or video shots respectively.  Codewords are then defined as the centers of the learned clusters. For the classification task, the latent topic with the highest probability is chosen as the category label of the image. It is worth to stress that during training the image/video categories are known, but the intermediate topic representation, used for testing are learn without additional supervision.

Beside image and video classification and segmentation, topic models have been largely used in the context of image retrieval, but mostly to textual information associated to image data (\cite{boyd2017applications}). For instance, in the context of the MediaEval Retrieving Diverse Social
Images Task was required participants to provide the most diverse and relevant images given a search query.  \cite{dudy2015ohsu} proposed to transform image tags and textual description features into a weighted term frequency-inverse document frequency bag-of-words representation (\cite{church1999inverse}), on the top of which they performed Latent Dirichlet Allocation (LDA) (\cite{blei2003latent}), to represent topic groups within the results.

A multimodal  approach to capture topics in massive social media data has been addressed by \cite{qian2016multi}. The model, based on the Multimodal-LDA is able to learn correlations between visual and textual modalities.

Unlike classical approaches, where topics are used as an intermediate and more powerful representation for classification that in turn is performed in a supervised manner, in our work topics are the result of a first layer of classification, obtained in fully unsupervised fashion. Furthermore, in all these models the discovery/classification results are given by the topic model itself and none of them uses topic models to drive a more detailed classification as in our system.

In the next section, we detail our proposed approach that provides an automatic hierarchical organization of smartphone pictures such the one shown in Fig.\ref{fig:approach} by relying solely on visual properties of images. As will be clarified in the next section, the number of topics, their names and the specific topic-related categories have been carefully chosen to address the problem of smartphone picture organization.

\section{Proposed approach}
\label{approach}

Our approach consists of two main steps: topic estimation and topic-related category classification.

\subsection{Estimating photo dominant topics}
To estimate the dominant topic in an image, we leverage a topic discovery method, called probabilistic Latent Semantic Analysis (pLSA) that has given excellent results in the field of document analysis  (\cite{hofmann2001unsupervised}).
Given a corpus of $N$ documents containing words from a vocabulary of size $M$, we would like to organize them in $K$ topics.

The  corpus of documents is summarized by a $M \times N$ co-occurrence matrix, where each element $X(w_i, d_j)$ with $i = 1,...,M$, $j = 1,...,N$ stores the number of occurrence of the word $w_i$ in document $d_j$.
In addition there is a latent variable $z_k$ associated with each occurrence of a word $w_i$ on a document $d_j$, that represents the topic. The goal of pLSA is to find the topic-specific word distribution $P(w|z)$ and the corresponding document-specific mixing proportions $P(z|d_j)$ which makes up the document specific word distribution $P(w|d_j)$. Formally,

\begin{equation}
P(w|d) = \sum_{k=1}^K P(z_k|d)P(w|z_k)
\end{equation}

pLSA assumes each document $d_j$ (with word vector $w$) to be generated from all
topics, with document-specific topic weights.
The model expresses each document as a convex combination of topic vectors in the latent space with mixture coefficients $P(z_k,d_j)$ for each document $d_j$, where $k \in \{1,...,K\}$. The topic vectors are common to all documents in the corpus and the mixture vectors are specific to each document. 
For example, in Tab. \ref{fig:topic_distribution} are shown the mixed coefficients of six words and it can be appreciated how most of these words have the highest coefficient in correspondence of the same topic since it is very likely to find them in the same paragraph of a document.

To learn the topic specific distribution $P(w|z$) all documents that constitute the training set  are pooled together and the PLSA model is fitted to the ensemble of documents for a specified number of topics. In particular, the Expectation Maximization (EM) algorithm (\cite{zhang2001segmentation}) is used to estimate the parameters $P(z)$, $P(w|z)$ and $P(z|d)$ that maximize the posterior probability $P(z|d,w)$.

\paragraph{Inference and classification}
Let us suppose that we are given an unseen document, $d_{test}$ and we would like to assign  a topic to it.
Given the distribution of words in the documents of the test set, say $P(w|d_{test})$,  the document specific mixing coefficients, $P(z|d_{test})$ can be computed using the so called \textit{folding-in} heuristic (\cite{hofmann2017probabilistic}). When we have a new document $d_{test}$, the EM algorithm is re-run, but this time the  topic-specific word distributions $P(w|z_k)$ are kept fixed to their previous values computed at training, while only the $P(z_k|d_{test})$ are updated. In this way, we obtain the mixed coefficient $P(z|d_{test})$ for the unseen document.
The $i-th$ document of test is assigned to the topic $k$ that maximizes the probability of the $k-th$ topic: 
\begin{equation}
 \argmax_k P(z_k| d_{test}^i), k=1,...,K. 
 \label{eq:inference}
\end{equation}





\begin{figure}[t!]
\centering
\includegraphics[width=0.5\textwidth]{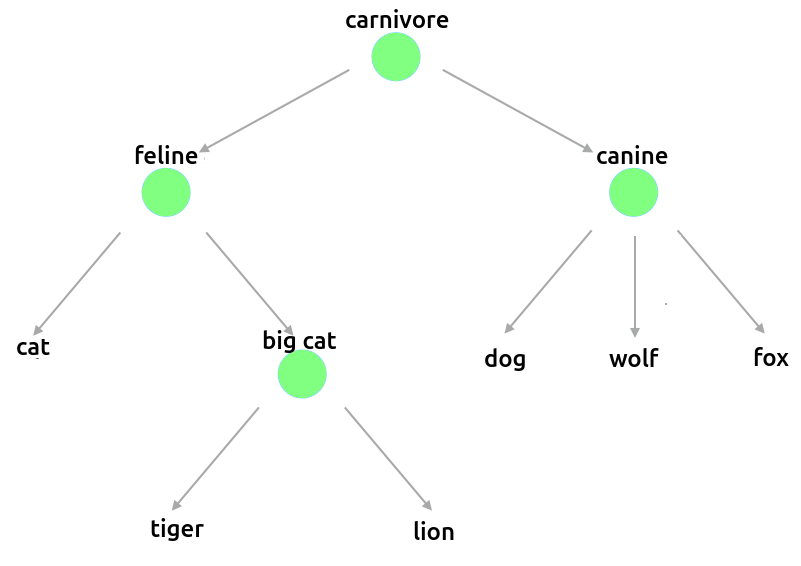}
\caption{WordNet directed acyclic graph. Each node corresponds to a synset and directed edge from node $u$ to node $v$ indicates that $u$ is an ancestor of $v$.}
\label{fig:dag}
\end{figure}

\paragraph{Translation in the image domain}

To adapt this framework to our context, we consider each image as a  \textit{document} and each tag, object in the image or concept describing the image as a  \textit{word} obtained by applying a concept detector or an object detection algorithm (\cite{yang2016exploit,redmon2016you}). In order to apply pLSA,  we need first to define a finite
vocabulary of words. We build the vocabulary starting by listing
all tags that have been used more than 5 times in the tranining set. This heuristics enforces that all rarely used tags are neglected.  If the tag appears only on a few personal photo collections,  it is considered rarely used,  independently
on the count.

\paragraph{Automated topic naming}
As a result of the inference, we obtain the mixture coefficients that allow to compute the dominant topic of an image with equation (\ref{eq:inference}). We automatically assign a name to the inferred topic $k$, by using the semantic similarity
between the top $Q$ words, (we took $Q=10$), defining the topic $k$ with highest confidence and $K=8$ predefined topic names, that we will denote by capitalized words hereafter, namely: \textit{Interior and Objects, Pets and Animals, Nature and Landscape, Food and Drinks, Street-view and Architecture, People and Portraits, Sport and Adventure,  Text and Visual}. 
The choice of these topics was inspired by the categories of Eden Photos and motivated by the need of having a small number of categories that could cover all possible content of smartphone pictures.
To compute this
semantic similarity, we leverage WordNet, a lexical database that groups English words into sets of cognitive synonyms, called \textit{synsets} (\cite{miller1995wordnet}).  All synsets are connected to other synsets by means of semantic relations.  Each vertex $v$ is an integer that represents a synset, and each directed edge $(u,v)$ represents that $u$ is a hypernym (ancestor) of $v$. The graph is directed and acyclic (see Fig.\ref{fig:dag}). We measure the semantic similarity between two words based on the shortest path in the hypernym taxonomy. Specifically, we used the
Lin function (\cite{lin1998information,meng2013review}), which is an Information Content (IC)-based similarity measure that relies on the most specific ancestor node, called Lowest Common Subsumer
(LCS).  Semantically, the LCS represents the commonality of the pair of concepts. For example, the LCS of {\em mosquito} and {\em bee} in WordNet is {\em insect}. If there are multiple candidates for the LCS (due to multiple inheritance), the LCS that results in the shortest path between two input concepts is chosen. Given two synsets, say  $s_1$ and $s_2$, their similarity is computed as, 
\begin{equation}
S(s_1,s_2)= \frac{2 \cdot IC(LCS(s_1,s_2))}{ IC(s_1) + IC(s_2)}.
\end{equation}
where $IC$ is a measure of specificity for a concept. Higher values are associated with
more specific concepts (e.g., chair), while those
with lower values are associated to more general concepts (e.g., doctrine).
In this work, the IC was derived from \textit{SemCor} (\cite{miller1993semantic}), a manually sense-tagged subset of the Brown Corpus (\cite{francis1982frequency}).  


We compute the sum of the Lin similarity between each of the 10 top tags defining the image and the two
words in the topic name, for each of the $K$ topics.
The topic that has the highest probability is the one that will
get assigned to the nameless topic, namely,

\begin{equation}
 \argmax_k  \sum_{i=1,..,Q,j=1,2} S(s_i,s_j^k)   , k=1,...,K 
\end{equation}
where $s_i$ is the synset associated to a tag of the image and $s_j^k$ is the synset associated to one of the two words defining the topic $k$.

\subsection{Estimating the topic-related categories}

After assigning a topic name to each picture, the proposed method provides a more detailed classification into topic-related category.

For each of our eight topics, we defined the corresponding topic-related categories by relying on topic-related largely used datasets whenever possible.
For example, for the topics \textit{Street-view and Architecture} and \textit{Nature and Landscape}, we used the categories of the Places dataset (\cite{zhou2017places}), that contains 10,624,928 images from 434 categories. We ended up with 277 categories for  \textit{Street-view and Architecture} and 88 categories for \textit{Nature and Landscape} respectively. 
For the \textit{Food and Drinks} topic, we used all 101 categories of the Food101 dataset (\cite{bossard2014food}).
For \textit{Sport and Adventure},  we used the categories of the UCF Sports Action Dataset (\cite{rodriguez2008action}) more those relate to Sport of the WIDER dataset (\cite{ahsan2017complex}). 
For \textit{Interior and Object} and \textit{Animal and Pets}, we manually selected the appropriate categories from the ImageNet dataset (\cite{deng2009imagenet}) and the Places dataset (\cite{zhou2017places}). This left us with 428 categories for Interior and Objects 
and 398 categories for Animals and Pets.
For \textit{Text and Visuals}, we defined the categories by inspecting a large training collection of photos captured by a smartphone and identifying images which contained text or some kind of artistic work. Getting specific categories defined is complicated, as many of these categories are defined by the context of the photo instead of the actual content. For example, what differentiates a recipe from class notes is the context and meaning of the text, rather than the visual features which define the image. With this in mind, we defined eleven visual categories which are as follows: \textit{map}, \textit{screen shot}, \textit{magazines}, \textit{drawing}, \textit{sign}, \textit{tattoo}, \textit{poster}, \textit{graffiti}, \textit{painting}, \textit{receipt}, \textit{writing}. Finally,  for \textit{Parties and People}, we defined the following eight categories: \textit{adult}, \textit{child}, \textit{selfie}, \textit{group}, \textit{family}, \textit{portrait}, \textit{manifestation}, \textit{conference} in addition to 5 categories of the PEC dataset (\cite{bossard2013event}): \textit{birthday}, \textit{concert}, \textit{exhibition}, \textit{graduation}, \textit{wedding}. The total number of categories for each topic are detailed on Table \ref{tab:appComparison}.

\section{Experimental results}
\label{experiments}
In this section, we  detail our experimental setting and the experiments performed. Then, we analyze and discuss the results. 


 
\subsection{Experimental setting}
\label{experiments_1}
\subsubsection{Dataset}
The training dataset was collected with the goal of covering the eight topics defined above. With this goal, we gathered personal photos taken by a mobile phone or a digital camera from 13 subjects having different hobbies (trekking, cooking, traveling, etc), for a total number of 13,845 images, with an average of 1,065 pictures per user. On Table \ref{tab:dataset_training}, the number of images per user and the number of different topics observed in the pictures are reported. 

The test dataset consists of a set of personal photos taken by a mobile phone belonging to 40 subjects, different from those who participated in the collection of the training set, for a total number of 14421 images, with an average of 360 pictures per user.

\begin{table*}[!h]
\centering
\caption{Topic names and number of categories per topics}
\resizebox{0.7\textwidth}{!}{%
\begin{tabular}{ |c | c | c |c | }
\hline 
  \multicolumn{2}{|c|}{\textbf{Eden photos}} &  \multicolumn{2}{|c|}{\textbf{Hierarchical photo organization}} \\ 
 \hline
 \textbf{Topic} & \textbf{\#classes} &  \textbf{Topic} & \textbf{\#classes} \\ \hline   Street-view  and Architecture & 1& Street-view and Architecture & 227\\ \hline  Nature and Landscapes  &1 &  Nature and Landscapes  & 88 \\ \hline   People and Portraits   & 1&  People and Portraits   & 6\\ \hline    Food and Drinks    & 1&  Food and Drinks & 101\\ \hline    Text and Visual   & 1&  Text and Visual & 11\\ \hline     Animals and Pets  &1 &   Animals and Pets & 398\\ \hline   Interior and Objects  & 1&  Interior and Objects &428 \\ \hline    Sports and Adventure &1 & Sports and Adventure & 40\\ \hline     Cars and  Vehicles & 1& Social events and Parties & 12\\ \hline     Macro and Flowers  &1 &  Null &  1\\ \hline    Sunrises and Sunsets & 1&  \multicolumn{2}{|c|}{\textbf{Google Photos}}    \\ \hline     Paintings \& Art & 1& \textbf{Topic}   & \textbf{\#classes} \\ \hline     Beaches and  Seaside & 1& Things  &  1100\\ \hline     Events and Parties & 1& Places  & Undefined
 \\ \hline \noalign{\vskip 0.25cm}   
\end{tabular}
}
\label{tab:appComparison}
\end{table*}

\begin{table*}[t]
\caption{Training dataset composition}
\centering
\begin{tabular}{c c c c c c c c c c c c c c}
\hline
 Subject ID \vline & 1 & 2 & 3 & 4 & 5 & 6 & 7 & 8 & 9 & 10 & 11  & 12 & 13 \\ 
\hline\hline
\# Images  & 527 & 499 & 3551 & 1000 & 1000 & 1000 & 729 & 1000 & 1200 & 823 & 502  & 827 & 1200 \\ 
\# Topics  & 2 & 3 & 5 & 2 & 5 & 3 & 4 & 3 & 2 & 3 & 4 & 3 & 1 \\
\hline
\end{tabular}
\label{tab:dataset_training}
\end{table*}


\begin{figure*}[t!]
\centering
\begin{tabular}{c r r r}
\includegraphics[width=0.8\textwidth]{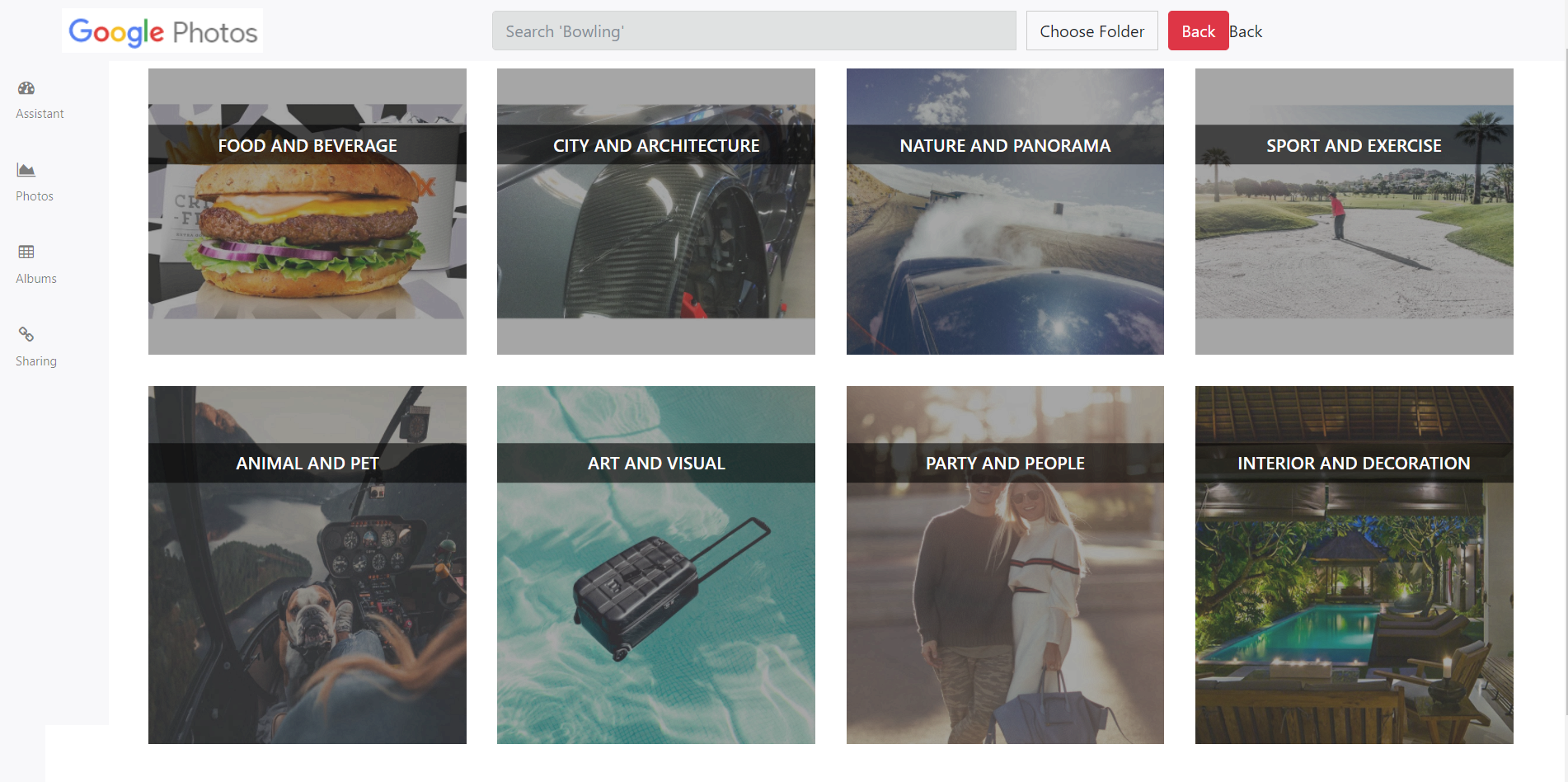}\\
\includegraphics[width=0.8\textwidth]{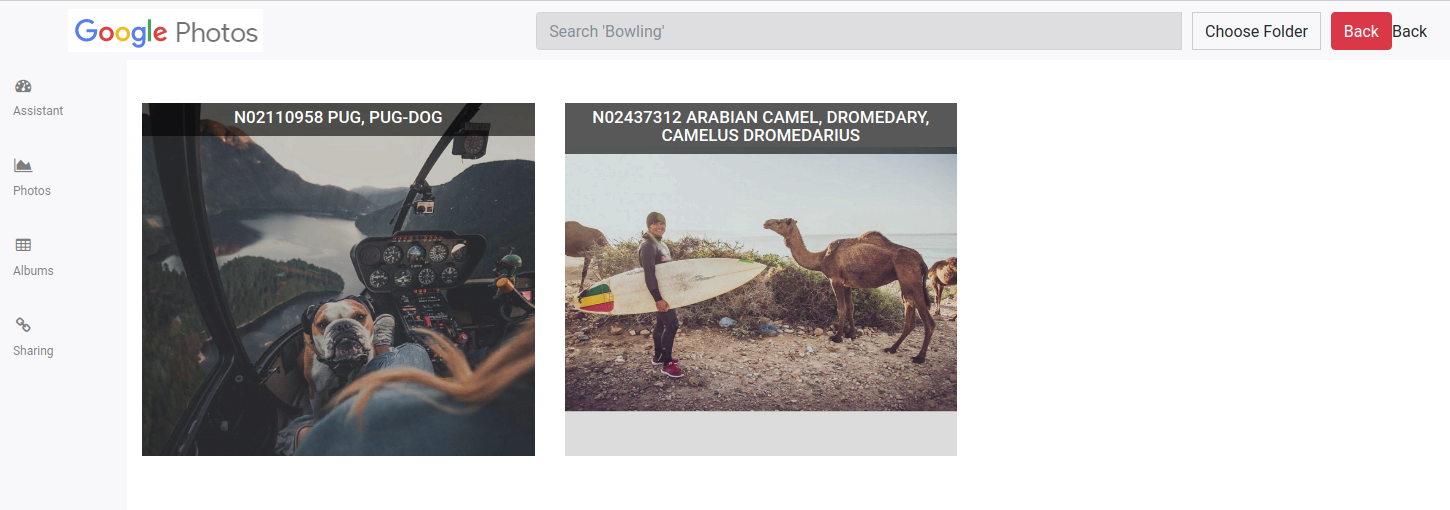}\\
\includegraphics[width=0.8\textwidth]{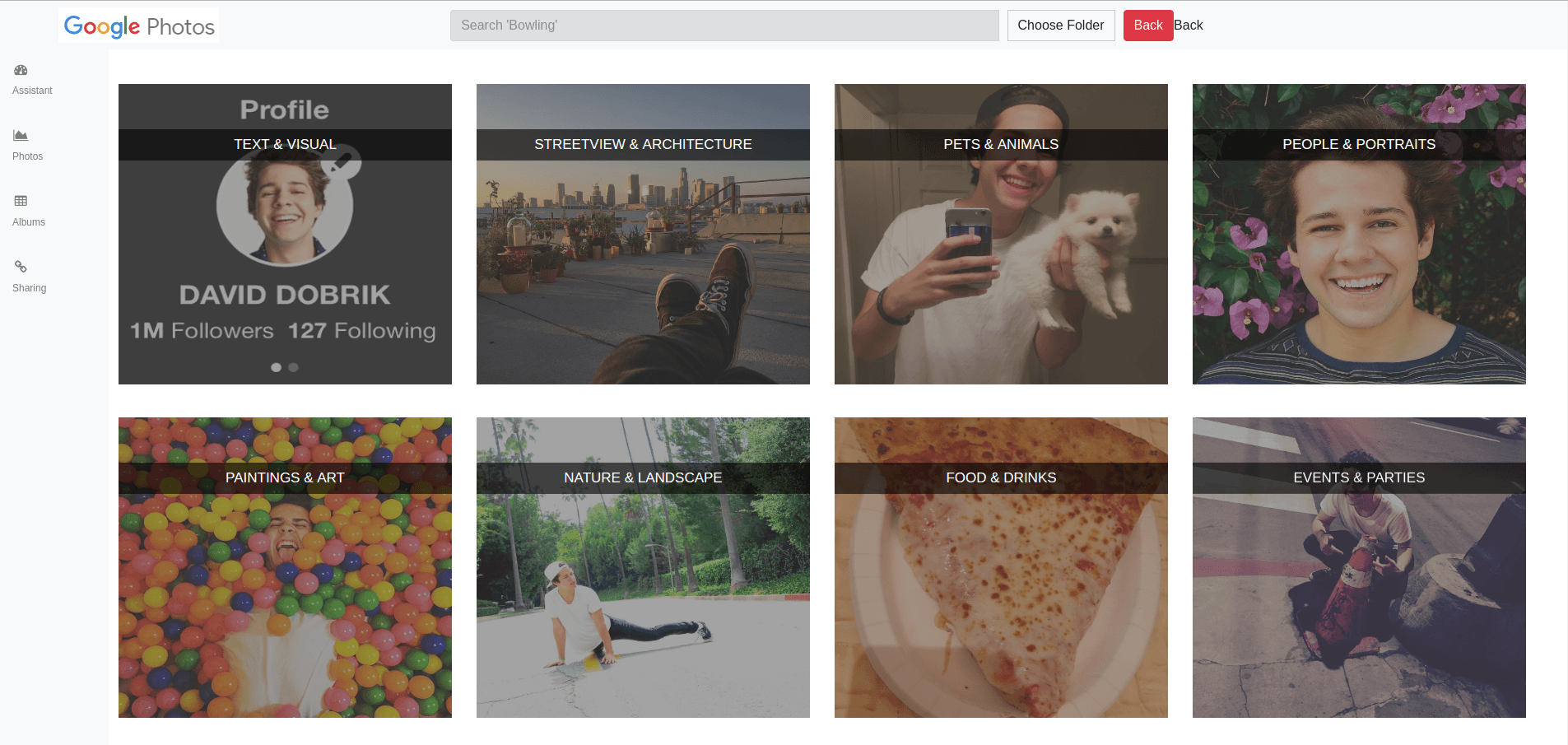}
\end{tabular}
\caption{Visual interface used to show to the participants of the user study the results of two
different systems. The top two images correspond to the results obtained with our system. In particular, the first image show the categories, and the second one the subcategories of \textit{Nature and Panorama}. The bottom image to the results obtained with
Eden photos.}
\label{fig:interface}
\end{figure*}

\subsubsection{Validation protocol}
\label{subsub_validation}
We evaluated three different aspects of our proposed approach: 
1) how good is the unsupervised classification into topics;  2) how much the users appreciate the proposed hierarchical organization and the appropriateness of the topic and topic-related categories; and
3) the overall classification accuracy of the system.

\paragraph{Topic coherence measures} To evaluate the performance of a topic model, several topic coherence measures have been proposed that take into account the average or median of pairwise word similarities formed by top words of a given topic. In this work, we used two widely used topic coherence measures: the UCI measure introduced by \cite{newman2010automatic} and the UMass measure introduced by \cite{mimno2011optimizing}. The UCI-score, $\mathcal{C}_{UCI}$ uses as pairwise score function, the Pointwise Mutual Information (PMI) and is defined as follows:
\begin{equation}
\mathcal{C}_{UCI} =\frac{2}{N(N-1)} \sum_{j=2}^N \sum_{i=1}^{j-1} \log{\frac{P(w_i,w_j) + \epsilon}{P(w_i)P(w_j)}},
\end{equation}
where $P(w_j, w_i )$ is the joint probability of ($w_i,w_j$) computed as the ratio
of number of documents containing both words $w_j, w_i$, $P(w_i)$ ($P(w_j)$) is the \textit{a priori} probability of $w_i$ ($w_j$) computed based on frequencies in the dataset, and $N$ is the total number of words. The smoothing count, $\epsilon$ is added to avoid calculating the logarithm of zero.


The UMass-score is also based on co-occurrences of word pairs, but measures how much, within the words used to describe a topic, a common word is in average a good predictor for a less common word.
More specifically, given an ordered list of words ordered by decreasing frequency $p(w|k)$, say $W = <w_1 , . . . , w_n>$, it is defined as:
\begin{equation}
 \mathcal{C}_{UMass} = \frac{2}{N(N-1)} \sum_{j=2}^N \sum_{i=1}^{j-1} log \frac{P(w_j, w_i) + \epsilon}{P(w_i)}.
\end{equation}
Note that the $\mathcal{C}_{UMass}$ has always a negative value.

Additionally, we report the average NMPI (annotated as $AvgNPMI$) among the top $Q$ words as an  internal measure of topic coherence:
\begin{equation}
AvgNPMI = \frac{1}{Q(Q-1)} \sum_{j=2}^N \sum_{i=1}^{j-1} \log \frac{P(w_j, w_{k})}{P(w_i, w_{k})}
\end{equation}
where $k$ indicates the k-th topic. The $AvgNPMI$ range is in the interval $\{ -1,1\}$.



\paragraph{Assessing the proposed organization through an user study}
The proposed approach has been evaluated through an user study, since ultimately the impact of the automatic organization depends on its value to the user. As subjects, we recruit both the 40 owners of the photo collections as well as 30
subjects not involved with the data collection in any way. The photo owners are a valuable resource to discern the photo organization quality, since they only have fully  experienced the original content.

We provided to all participants an Information Sheet that gave them the necessary understanding for the motivation and procedures of the study. To measure the quality of our organization on an absolute scale and to allow independent judges to evaluate the photo organization usefulness, we asked each owner to  provide  “ground-truth” categories of his/her pictures. Specifically, we asked the users to provide a list of categories that emphasizes the dominant topics in his/her pictures.
Then, the users are asked to compare each pair of systems, each one of the pair shown in a different browser tab. To make the systems blind to the users and to avoid bias judgment due to different visualization, we mimicked the visual interface of Google Photos and presented all results using the same interface (see Fig. \ref{fig:interface}). As shown in Fig. \ref{fig:interface}, the users first see the picture type, and then, by clicking on a picture type they can visualize the subcategories or the pictures corresponding to the picture type. In each subcategory folder, are shown the pictures inside that folder.
We asked to the participants two questions: The first question that evaluates aspect a), was: \textit{Which kind of organization do you prefer and why, independently on the accuracy?} The second question, that evaluates aspect b), was: \textit{Which system do provide more accurate results, independently on the organization?} 

\begin{table*}
\caption{Architecture and dataset used for pre-training for each topic}
\centering
\resizebox{0.6\textwidth}{!}{%
\begin{tabular}{c c c}
\hline\hline
Topic & Architecture & Dataset used for pre-training\\
\hline
Interior and Decoration & ResNet-50 & ImageNet\\[0.3ex]
Party and People & ResNet-50 & Places\\[0.3ex]
Art and Visual & ResNet-101 & ImageNet\\[0.3ex]
Animal and Pet & ResNet-50 & ImageNet\\[0.3ex]
Sport and Exercise & ResNet-101 & ImageNet\\[0.3ex]
Nature and Panorama & VGG-16 & Places365\\[0.3ex]
City and Architecture & VGG-16 & Places365\\[0.3ex]
Food and Beverage & ResNet-50 & Food-101\\[0.3ex]
\hline
\hline
\end{tabular}
}
\label{tab:cnn_used}
\end{table*}

\subsubsection{Experiments}

For each user in our test set, we first estimated to which topic the image belongs to and then we classified the image accordingly to the categories of the topic at hand. 
For instance, if the algorithm predicts that the image belongs to the topic {\em Animals and Pets}, than a more detailed classification of the pictures is performed with the classes {\em cats, dogs, births, horses, etc.}. In this work, we used a concept detector developed by Imagga Technologies Ltd. Imagga's auto-tagging technology\footnote{\url{http://www.imagga.com/solutions/auto-tagging.html}}. The advantage of Imagga's Auto Tagging API is that it can directly recognize over 2,700 different objects and in addition it returns more than 20,000  abstract concepts (corresponding to the words) related to the analyzed images. 
The total number of tags found in the training set is 13,852.   The number of tags after the filtering is 3,312.
We then applied pLSA to learn  the topic specific word distribution $P(w|z_k)$. At test time, we applied the folding-in heuristic detailed in section \ref{approach} keeping  $P(w|z_k)$ fixed and we obtained the mixture coefficient $P(z_k|d_{test})$. 
We automatically assigned a label to the topic with the largest probability. However, if the highest probability is below a given threshold ($0.035$ in our experiments), the picture is assigned to the \textit{Null} topic.

\begin{figure*}[th!]
\centering
\includegraphics[width=0.7\textwidth]{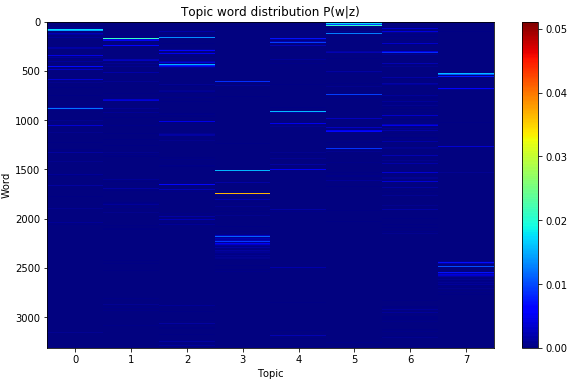}
\caption{Topic-specific word distributions $P(w|z_k)$ estimated from the training set.}
\label{fig:pwz} 
\end{figure*}

\begin{table*}[th]
\caption{Comparison of topic models (plSA, LDA, LSA) in terms of topic coherence measures}
\centering
\resizebox{\textwidth}{!}{%
\begin{tabular}{c c c c  | c c c  | c c c  }
\hline\hline
&  & \textbf{pLSA} & & & \textbf{LDA} & & &  \textbf{LSA} &\\
\hline\hline
Topic & UCI-score & Umass-score  & NPMI Topic & UCI-score & Umass-score  & NPMI Topic & UCI-score & Umass-score  & NPMI \\
\hline
Interior and Decoration & 1.40 &  -1.65 & 0.16 & - & - & - & -0.94 (-0.79)&  -2.80 (-2.60)  & -0.01 (0.00)\\[0.3ex]
Party and People & 1.54 & -1.72 & 0.16 & 0.87 & -0.80 & 0.20 & - & - & -\\[0.3ex]
Art and Visual & 1.60 & -1.58 & 0.20 & -1.10 (-4.63) &  -3.76 (-9.36) & -0.05 (0.03) & 3.39 &  -0.59 & 0.52\\[0.3ex]
Animal and Pet & 1.65 & -1.87 & 0.16 & -7.27 & -11.43 & -0.13 & 2.31 (1.98) &  -1.10 (-1.34)& 0.39 (0.30) \\[0.3ex]
Sport and Exercise & 1.41& -1.80 & 0.13 & - & - & - & -1.83 & -5.04  &  0.13\\[0.3ex]
Nature and Panorama & 1.76 & -1.69 & 0.21 & -5.28 & -9.38 & -0.05 & - & - & -\\[0.3ex]
City and Architecture & 1.41 & -1.69 & 0.15 & 1.06 &  -0.85 & 0.22 &  1.12 &  -0.58 (-5.04) & 0.23\\[0.3ex]
Food and Beverage & 1.44 & -1.78 & 0.14 & -1.12 (-4.10) & -0.19 (-3.88) & -0.05 (0.04) &  -0.37  & -0.12 & 0.06\\[0.3ex]
\hline
\textit{Average Topic Coherence}  & \textbf{1.53}  & \textbf{-1.72} & 0.16 & -2.69  & -4.95 & 0.02 & 0.6  & -2.23 & \textbf{0.20}\\[0.3ex]
& & & \\
\hline
\hline
\end{tabular}
}
\label{lab:label}
\end{table*}

As it can be appreciated in Table \ref{tab:appComparison}, our eight topic categories are a subset of the topic categories in Eden Photos. This is because our system allows several categories for each topic, so that the  \textit{Sunrises and Sunset}, \textit{Beaches and Seaside} and \textit{Flowers} can be considered as categories of \textit{Nature and Landscape} instead of being a topic themself. Similarly,  we treated \textit{Painting and Art} a category of  \textit{Text and Visual} and \textit{Cars and Vehicles} as a subcategory of  \textit{City and Architecture}. 

\begin{table*}[ht]
\caption{Results of the user study based comparison of our system vs Google Photos (top) and our system vs Eden Photos (bottom) on the dataset consisting of 40 users. Numbers indicate percentage of responses for each question. 
}
\centering
\resizebox{1.0\textwidth}{!}{%
\begin{tabular}{c c c c c c c c c c c}
\hline \hline
&  & Much better (5) & Better (4) & Similar (3) & Worse (2) & Much worse (1) & \textbf{Mean} & \textbf{Std} & \textbf{Up pvalue} & \textbf{ICC1}\\
\hline
\textbf{Photos} & Organization &  55\%& 42.5\% & 2.5\%& 0\% & 0\%  & 4.55 &  0.55 & 1 & -  \\[0.3ex]
\textbf{owners} & Accuracy & 0\% & 20\% & 40\% & 37.5\% & 2.5\% & 2.77 & 0.80 &  0.04 & -  \\[0.3ex]
\hline \hline
\textbf{External} & Organization &  48.34\%& 40.83\% & 10.00\%& 0.83\% & 0\% & 4.36 & 0.69 & 1 & 0.430\\[0.3ex]
\textbf{evaluators}& Accuracy & 1.66\% & 16.67\% & 47.5\% & 31.67\% & 2.5\% & 2.84 &  0.78 & 0.01 &  0.436\\[0.3ex]
\hline  \noalign{\vskip 0.35cm}
\hline
&  & Much better (5) & Better (4) & Similar (3) & Worse (2) & Much worse (1) & Mean & Std  Up pvalue & ICC1\\
\hline
\textbf{Photos} & Organization & 40\% & 55\% & 5\% & 0\% & 0\% & 4.35 &  0.57 & 1 & - \\[0.3ex]
 \textbf{owners} & Accuracy  & 0\% & 25\% & 55\% & 20\% & 0\% & 3.05 & 0.67 & 0.67 & -  \\[0.3ex]
\hline \hline
\textbf{External} & Organization &  19.66\%& 63.34\% & 16.67\%& 0.84\% & 0\% & 4.07 & 0.65 & 1 & 0.402\\[0.3ex]
\textbf{evaluators}& Accuracy & 4.16\% & 19.16\% & 50.84\% & 24.17\% & 1.67\% & 2.93 & 0.86 & 0.20 & 0.403\\[0.3ex]
\end{tabular}
}
\label{tab:user_study1}
\end{table*}
\begin{table*}[t]
\caption{Results of the user study based comparison of our system vs Google Photos (top) and our system vs Eden Photos (bottom) on the dataset consisting of 5 Instagram's vloggers. Numbers indicate percentage of responses for each question. The top rows report the evaluation made by the photo-owner, whereas the bottom rows refer to the average evaluation made by three users that saw the pictures for the first time.}
\centering
\begin{tabular}{c c c c c c c c c c}
\hline \hline
 & & Much better & Better & Similar & Worse & Much worse & &\\
\hline
\textbf{Google} & Organization &  55\% & 42\% & 3\%& 0\% & 0\% &  &\\[0.3ex]
\textbf{Photos} & Accuracy & 0\% & 20\% & 40\% & 37.5\% & 2.5\% & &\\[0.3ex]
\hline  \noalign{\vskip 0.35cm}
\hline
& & Much better & Better & Similar & Worse & Much worse & &\\
\hline
\textbf{Eden} & Organization & 41\% & 55\% & 14\% & 0\% & 0\% & & \\[0.3ex]
\textbf{Photos} & Accuracy  & 0\% & 30\% & 20\% & 50\% & 0\% & & \\[0.3ex]
\end{tabular}
\label{tab:user_study2}
\end{table*}

\subsection{Results and discussion}
\label{experiments_2}
In the following, we report and discuss the results obtained for topic discovery and assignment,
as well as the results of the user study.

Once classified into topics, the images were fed to the corresponding CNN that classified them into topic-related categories. A description of the CNN architectures used for each topic and the initial weights used are provided in Table \ref{tab:cnn_used}.  
In order to build the training dataset for fine-tuning, we needed a large amount of photos, ideally taken with a smartphone, as these are impromptu ones, that can be blurred or lacking proper lightening or having the motif of the photo off-centered.
With the goal of getting a large amount of smartphone pictures,  we scraped social media, such as Instagram and Flickr, and we also got additional photos from Google Images when needed. We automatically collected a large amount of photos per category, and later we manually filtered the ones that did not fit our criteria. Our goal was to get at least a thousand images per category to be able to fine-tune a pre-trained CNN.

\subsubsection{Topic discovery}

After fitting the pLSA model to our training set with 8 topics and automatically assigning a label to each word distribution, we obtained the following topic definitions:

\begin{itemize}
\item \textit{Food and Beverages}: fresh, healthy, dinner, eating, plate, meal, restaurant, delicious, diet, lunch, tasty, gourmet, snack, cuisine, nutrition, dish, vegetable, meat, cook, breakfast, pepper, sauce, tomato, vegetables, slice, kitchen, hot, cheese, bread, bowl.
\item \textit{Animals and Pets}: animal, dog, canine, domestic animal, pet, mammal, domestic, person, hunting dog, fur, cat, animals, funny, pets, adorable, sporting dog, purebred, terrier, feline, puppy, breed, hound, furry, kitten, eye, toy dog, spaniel, fluffy, little, whiskers. 
\item \textit{Art and Visual}: paper, element, shape, text, frame, drawing, money, card, flower, letter, blank, internet, representation, decorative, curve, currency, note, artistic, sketch, surface, document, book, floral, swirl, textured, leaf, information, creative, word, writing.
\item \textit{Nature and Panorama}: europe, rocks, shoreline, barrier, surf, seaside, asia, boundary, sunrise, hill, sunshine, seashore, ship, vessel, evening, sandbar, structure, rocky, peace, coastal, geological formation, turquoise, natural elevation, cloudscape, dusk, pacific, cliff, panorama, scenics, breakwater.
\item \textit{Parties and People}: person, caucasian, boy, couple, together, girls, clothing, indoors, family, friends,  teenager, 20s, two,  group, standing, friendship, working, laughing, blond, brunette, teen, student, romance, education, kid, adults, relationship, mother, romantic, healthy. 
\item \textit{Sport and Exercise}: person, caucasian, boy, active, healthy, family, kid, playing, play, athlete, childhood, ball, children, player, activity, team, game, two, exercise, little, training, soccer, football, baby, mother, fitness, toddler, match, adorable, care. 
\item \textit{City and Architecture}: group, crowd, spectator, town, pedestrian, buildings, stage, event, transportation, dark, vehicle, meadow, skyline, high, panorama, snow, music, center, aerial, wheeled vehicle, party, entertainment, tower, countryside, disco, club, transport, power, dance, concert.
\item \textit{Interior and Decoration}: wall, window, structure, interior, wood, door, furniture, luxury, estate, apartment, living, exterior, decor, sofa, residential, real, indoors, religion, comfortable, monument, tower, town, famous, church, inside, couch, lamp, historical, brick, europe.
\end{itemize}

\begin{figure*}[t!]
\centering
\begin{tabular}{ c c c c c }
\includegraphics[width=0.18\textwidth]{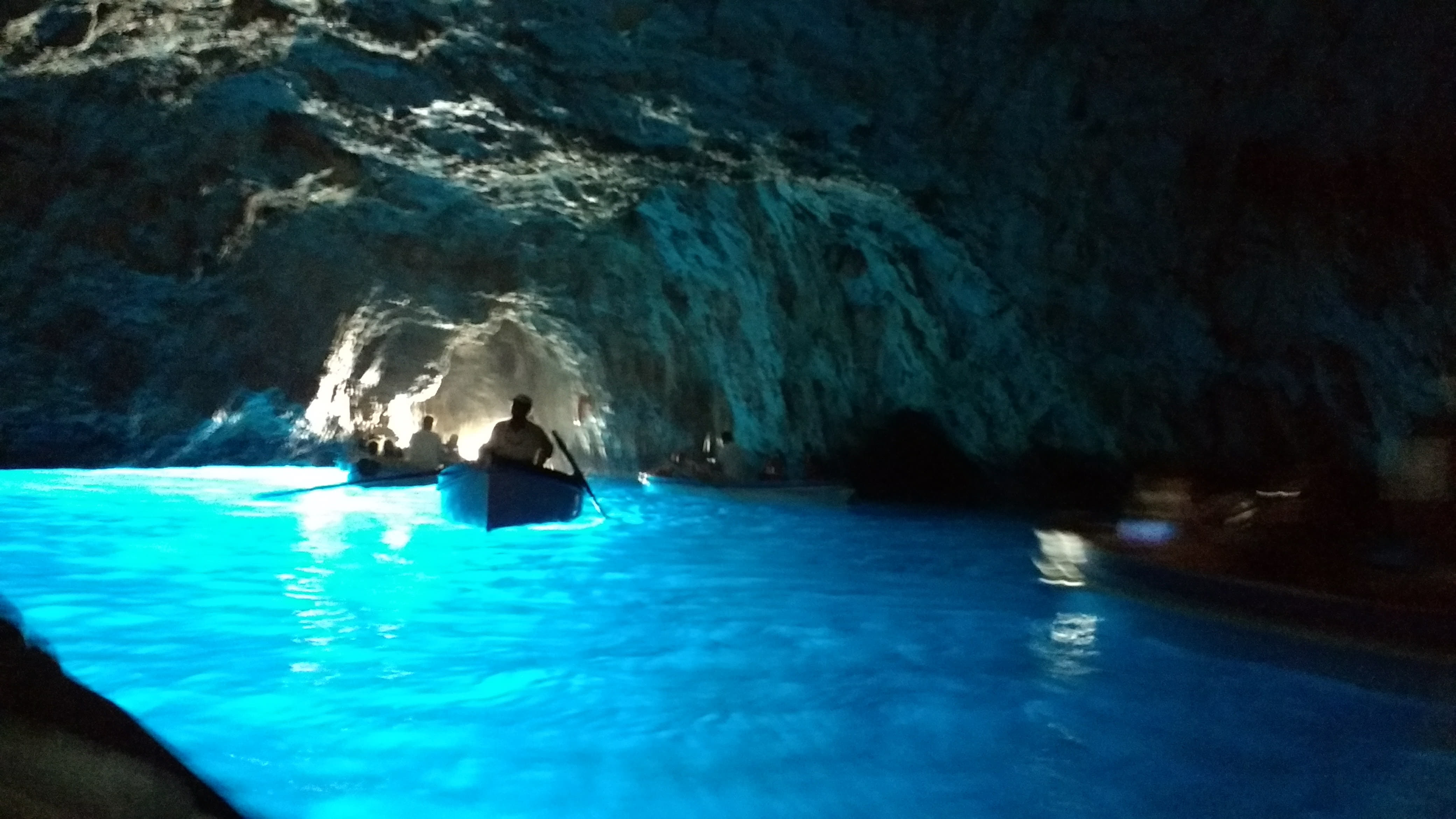}&
\includegraphics[width=0.18\textwidth]{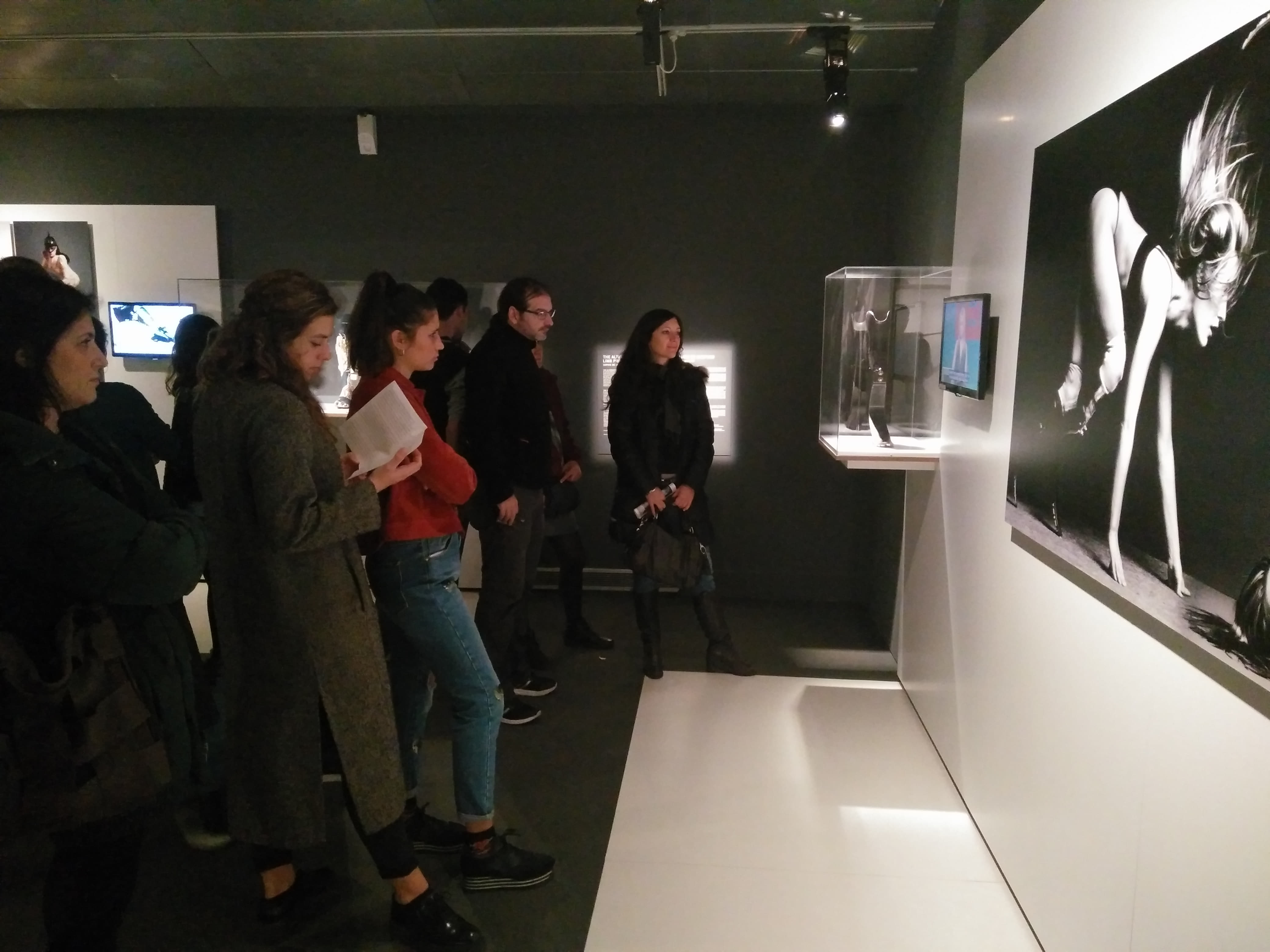}&
\includegraphics[width=0.18\textwidth]{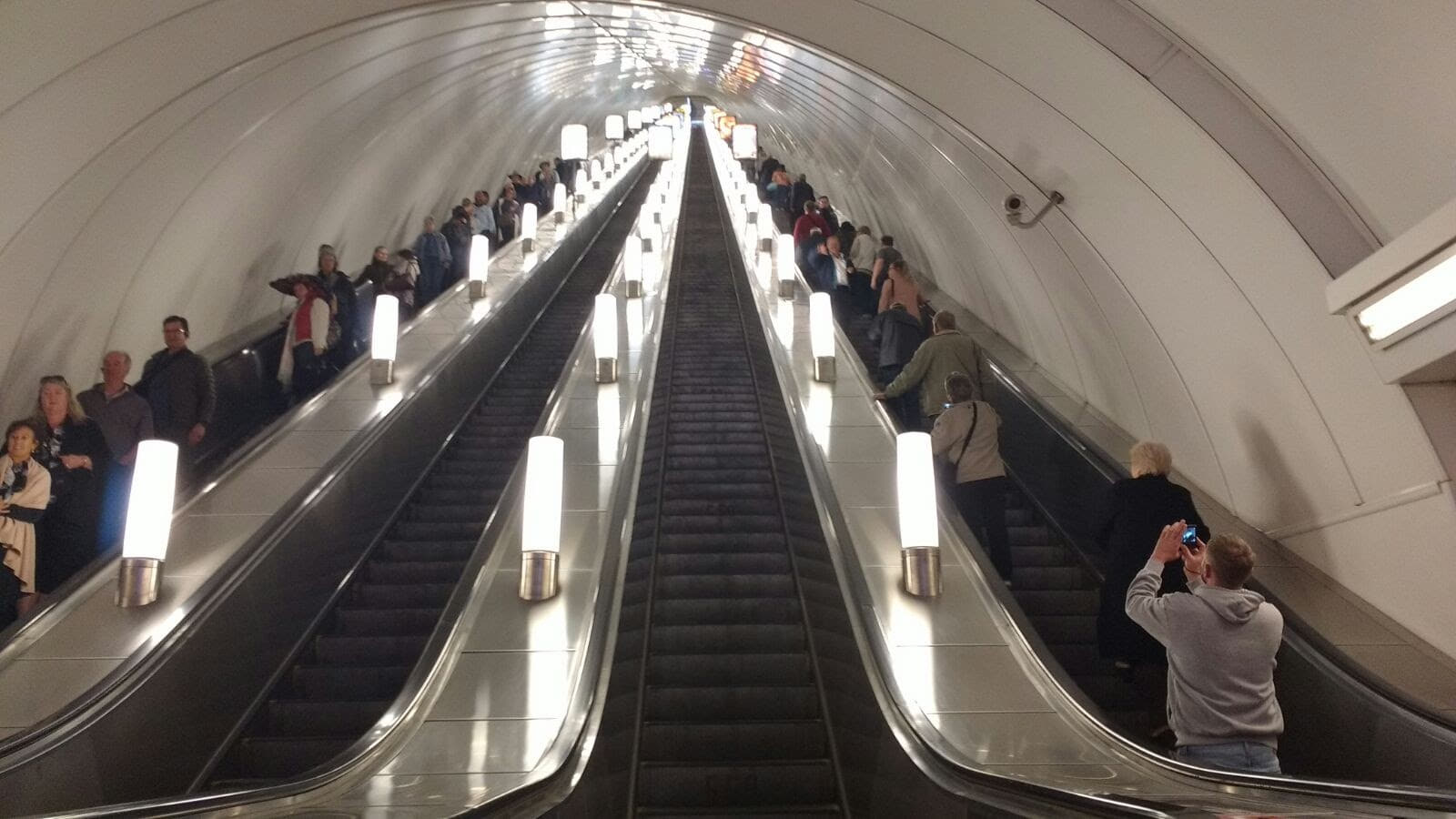}&
\includegraphics[width=0.18\textwidth]{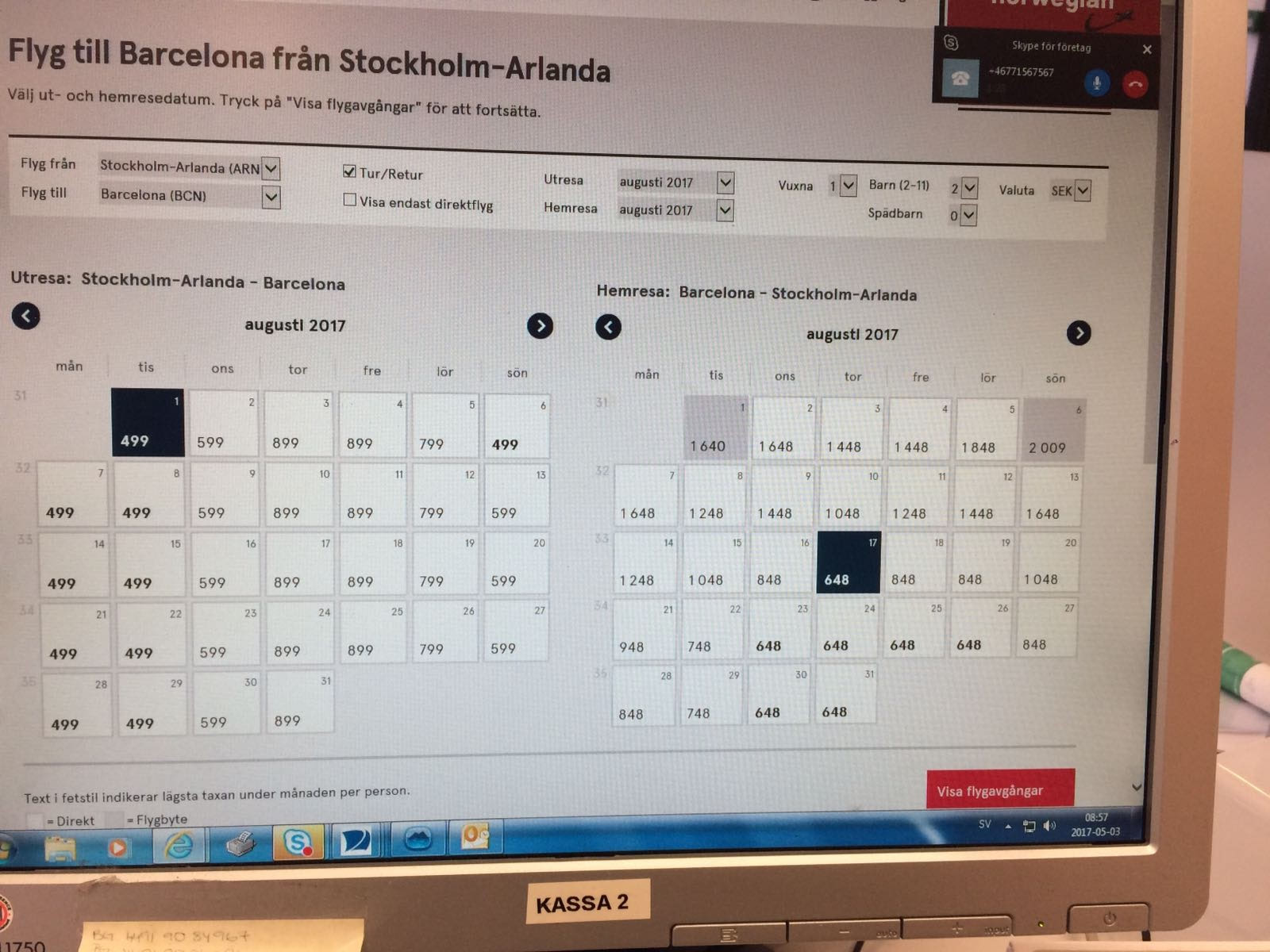}&
\includegraphics[width=0.18\textwidth]{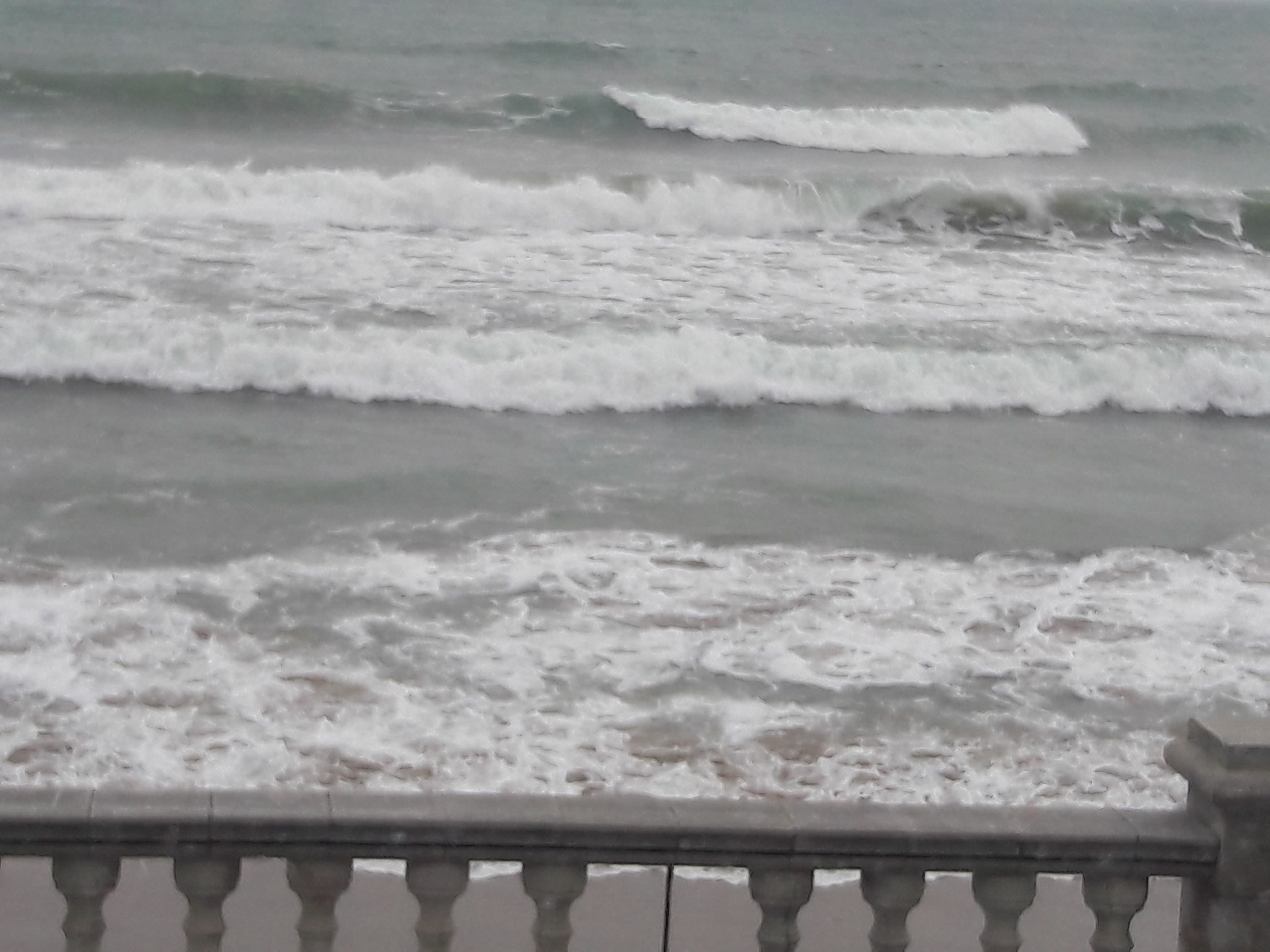}\\
grotto & exhibition & escalator indoor & screenshot & wave\\
\includegraphics[width=0.18\textwidth]{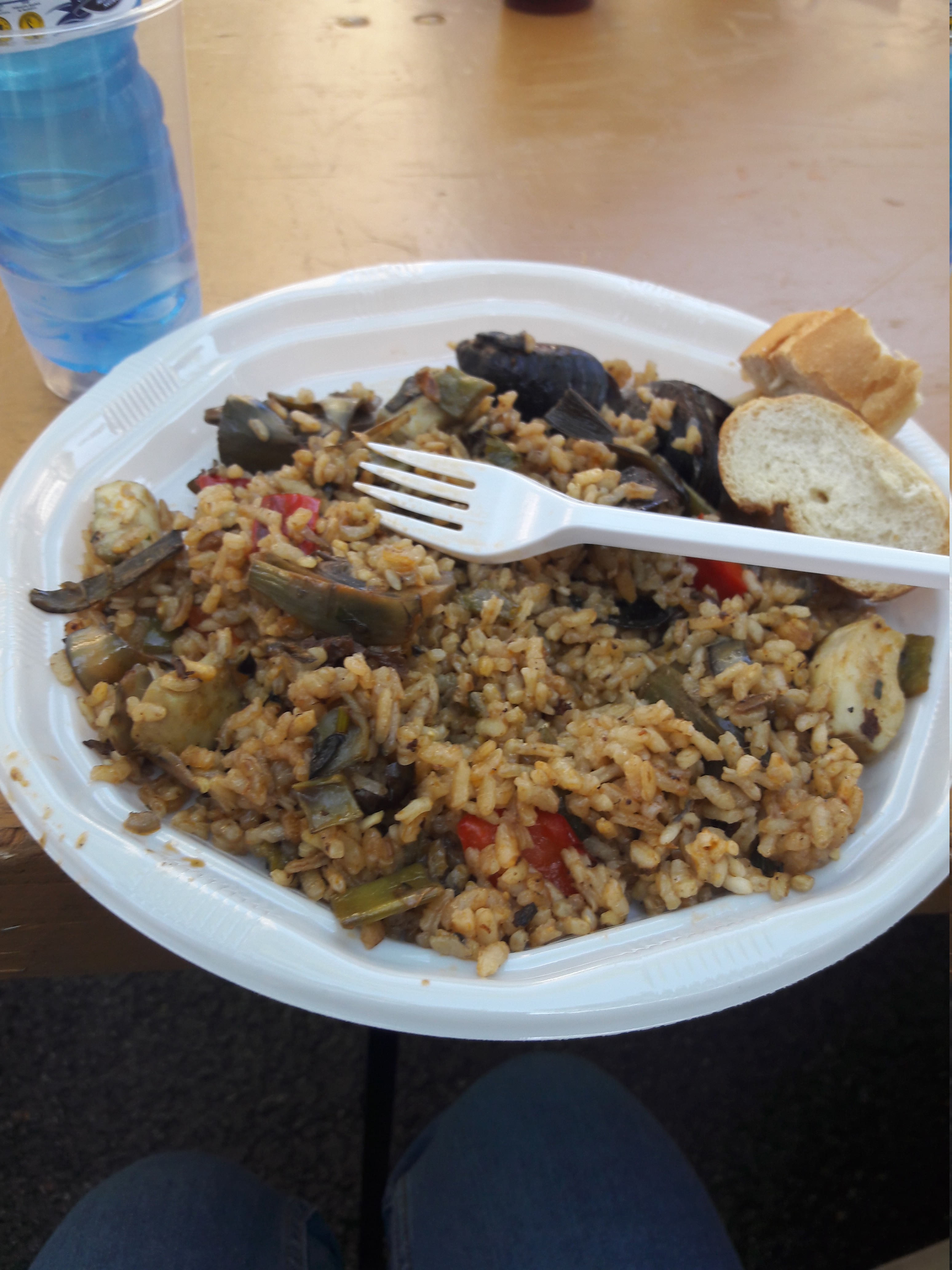}&
\includegraphics[width=0.18\textwidth]{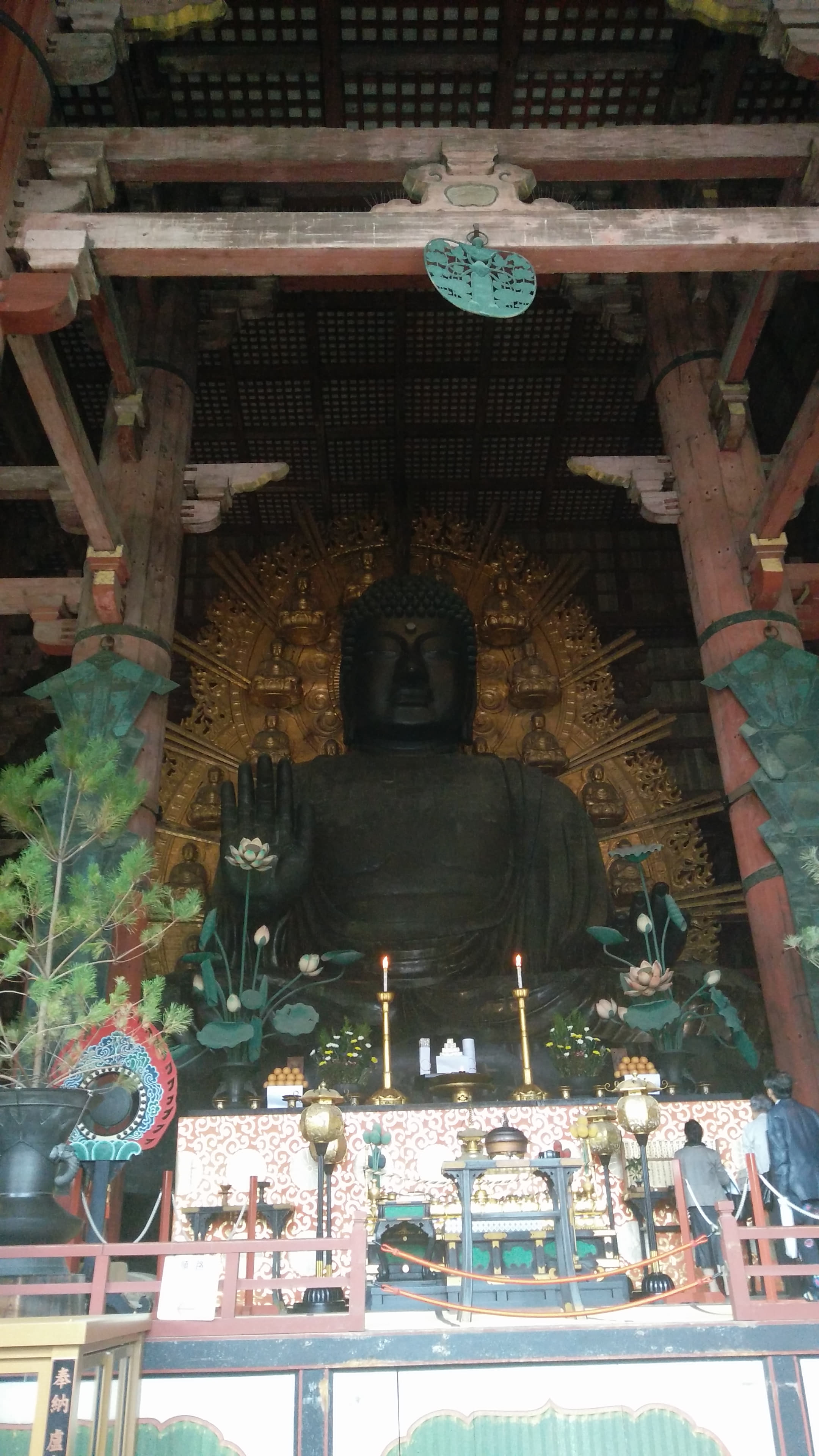}&
\includegraphics[width=0.18\textwidth]{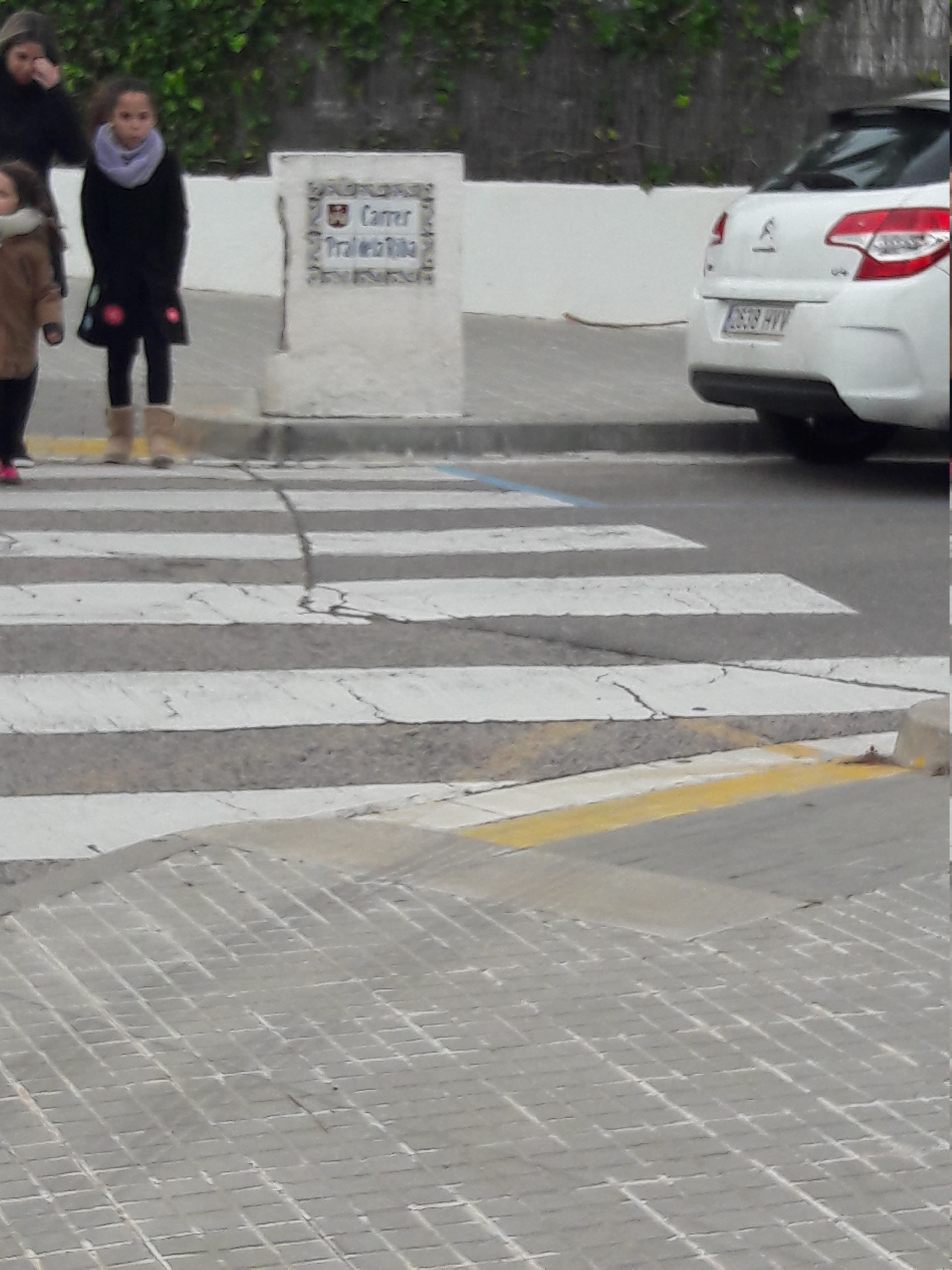}&
\includegraphics[width=0.18\textwidth]{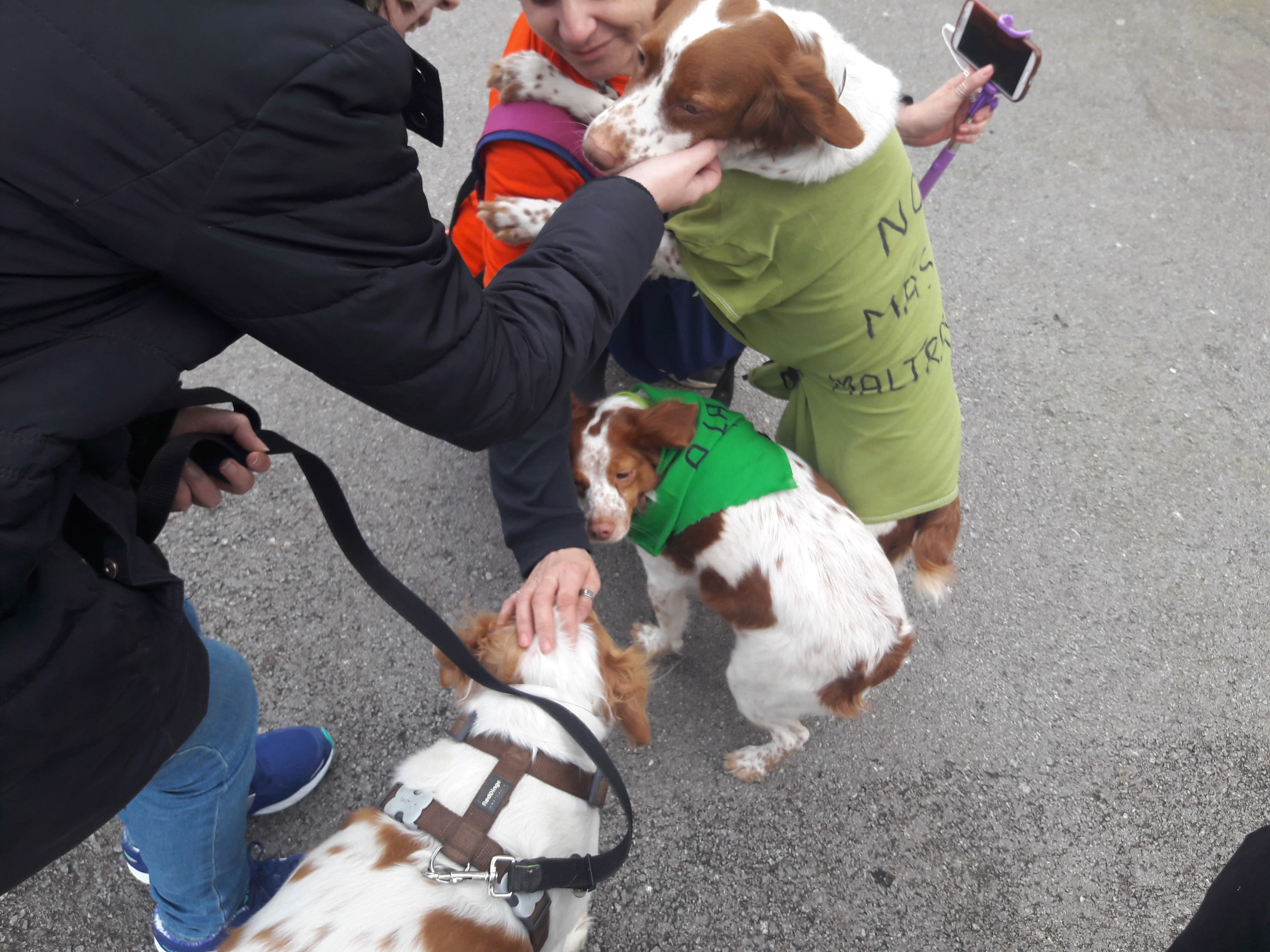}&
\includegraphics[width=0.18\textwidth]{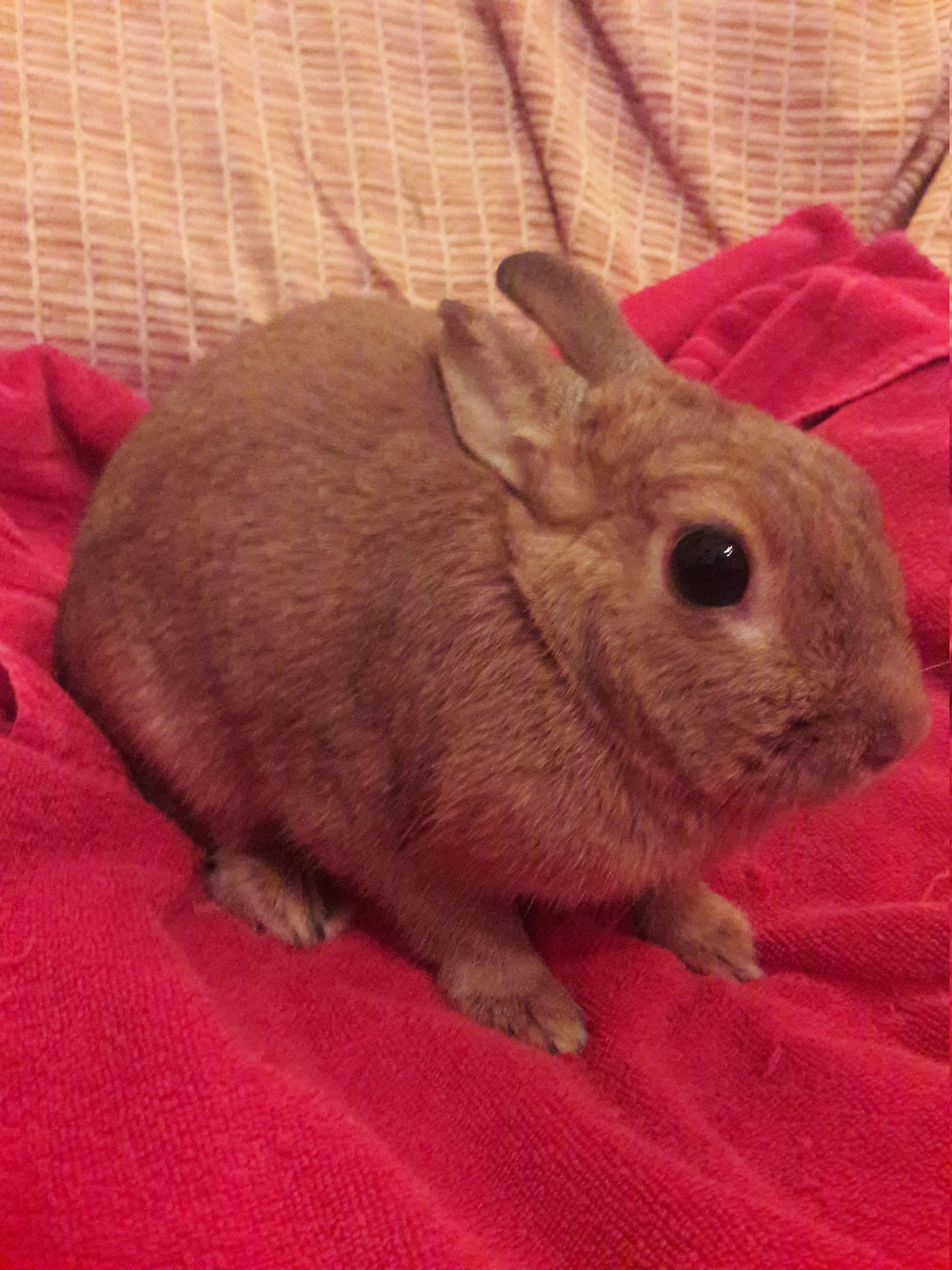}\\
paella & altar & crosswalk & brittany spaniel & hare\\
\end{tabular}
\caption{Example of images correctly classified by our system}
\label{fig:positive}
\end{figure*}

\begin{figure*}[t!]
\centering
\resizebox{1.0\textwidth}{!}{%
\begin{tabular}{c c c c c}
\includegraphics[width=0.2\textwidth]{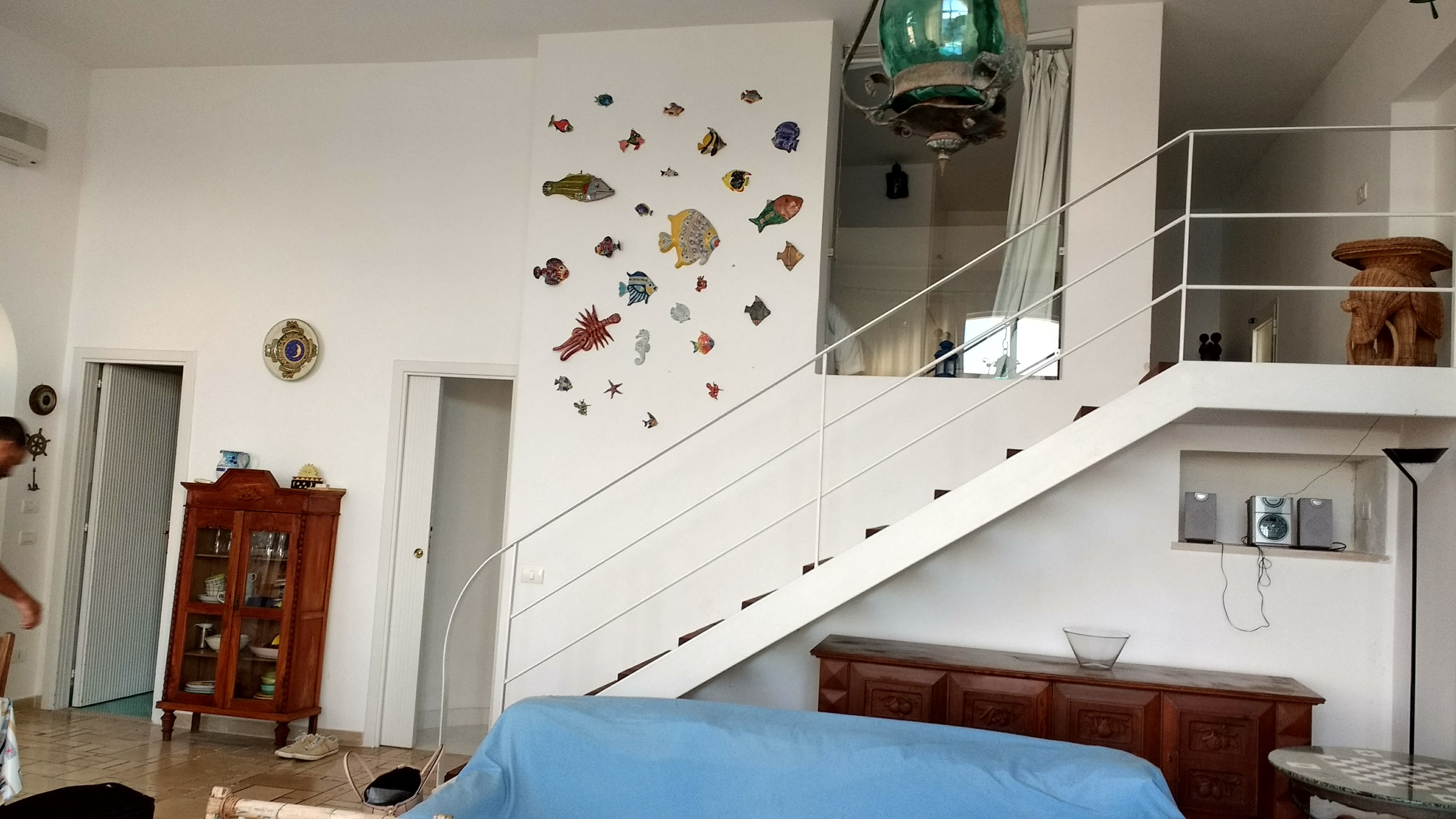}&
\includegraphics[width=0.2\textwidth]{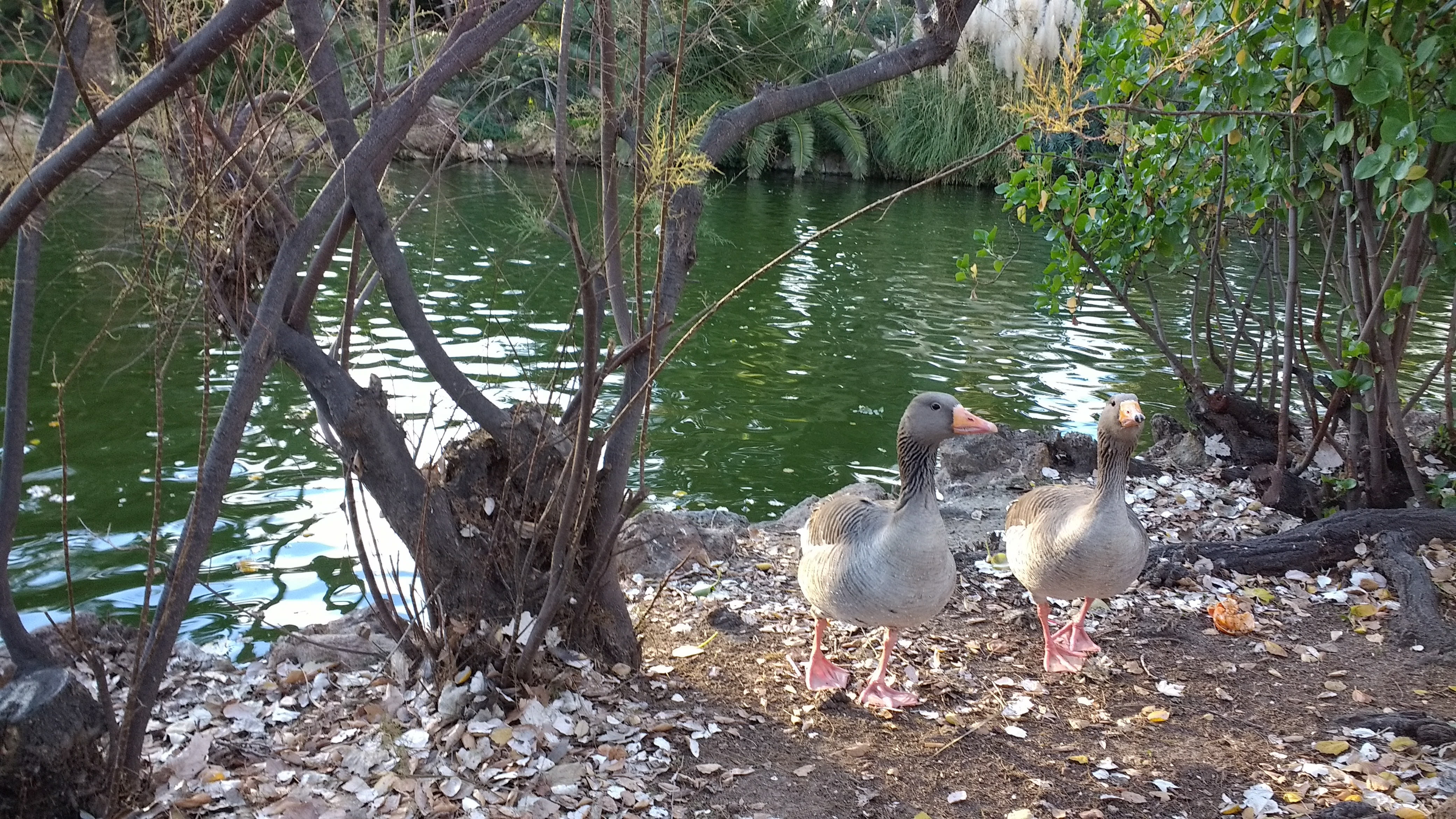}&
\includegraphics[width=0.2\textwidth]{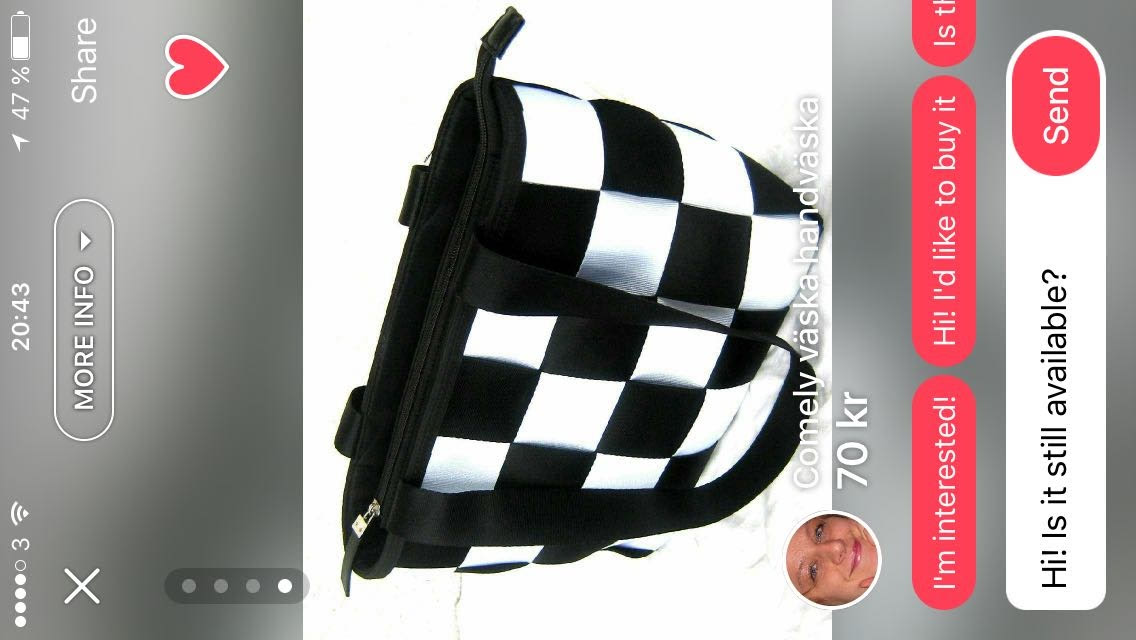}&
\includegraphics[width=0.2\textwidth]{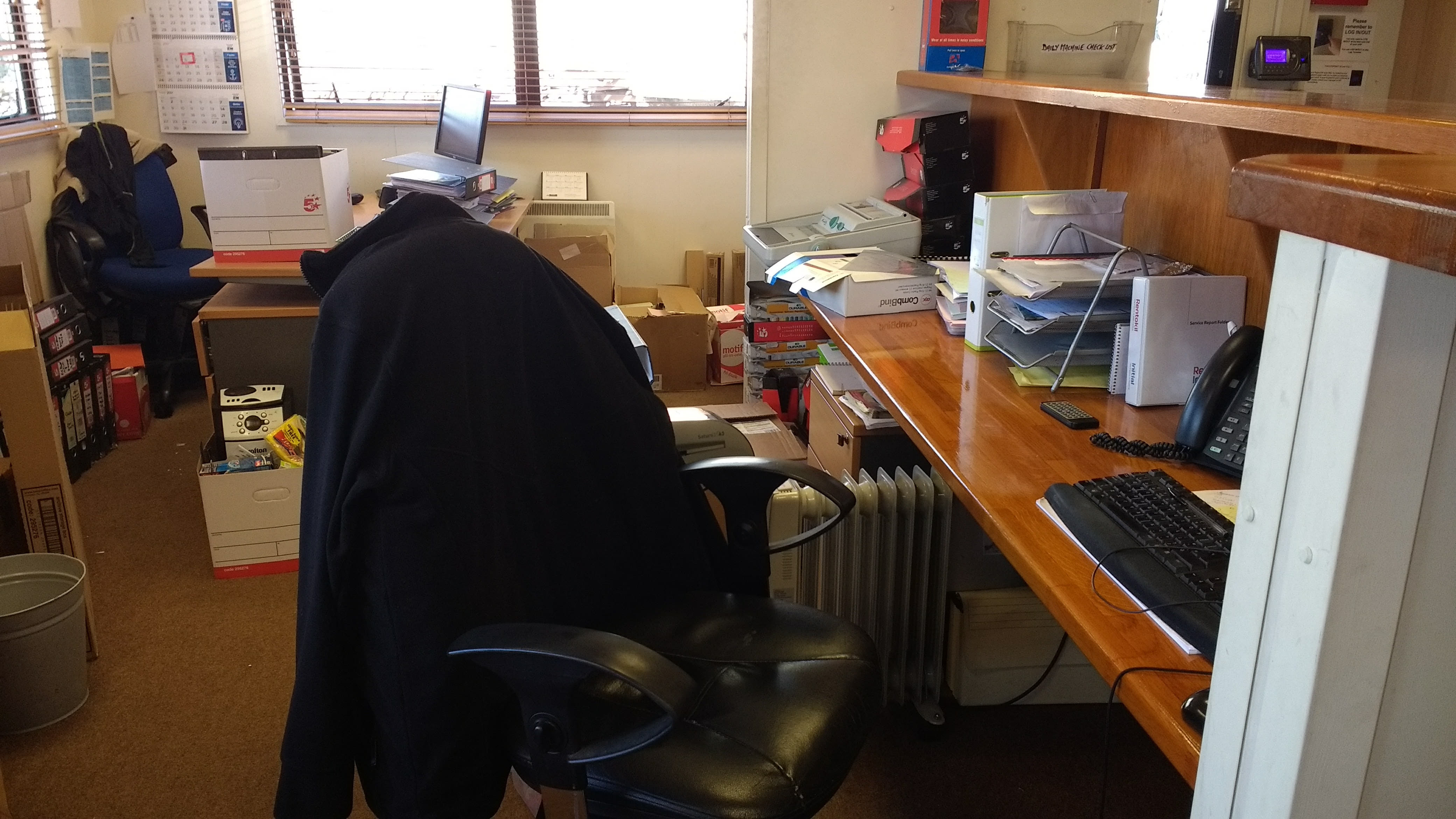}&
\includegraphics[width=0.2\textwidth]{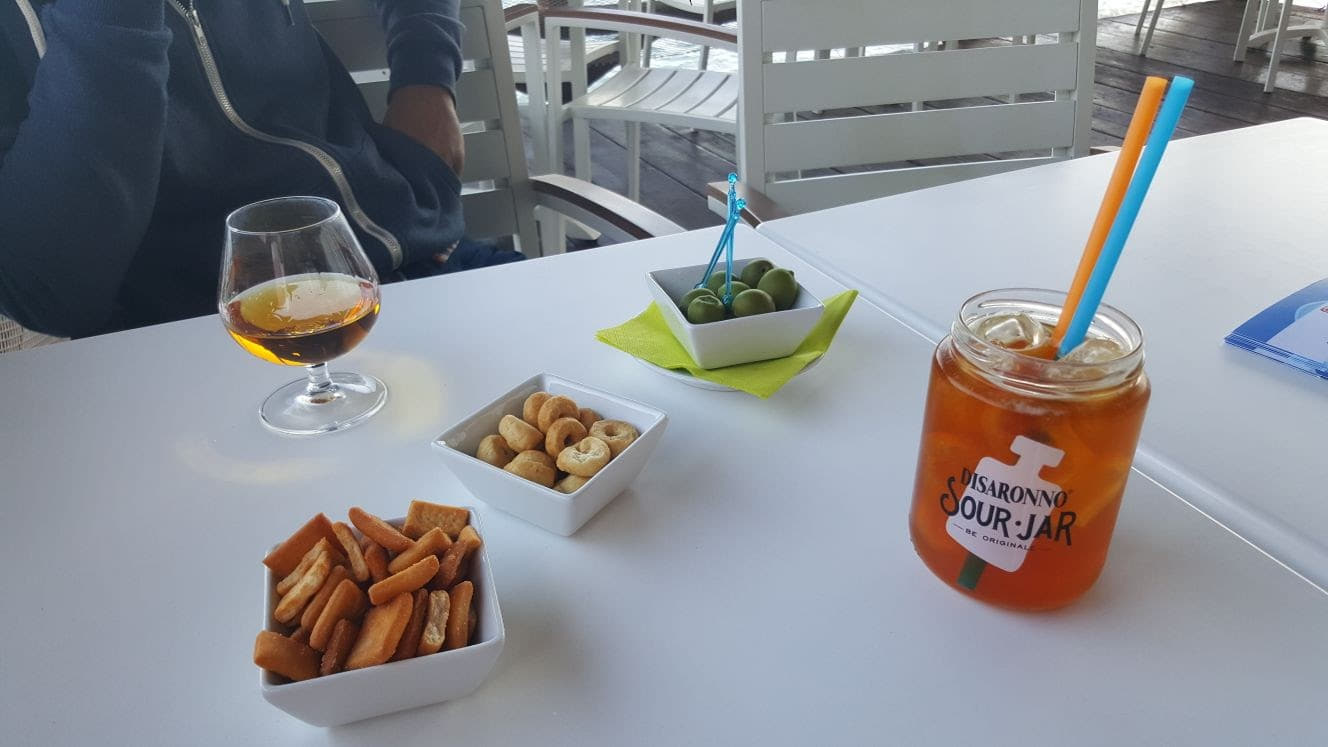}\\
bathtub & cottontail rabbit & poster & grand piano & hamburger\\
\includegraphics[width=0.2\textwidth]{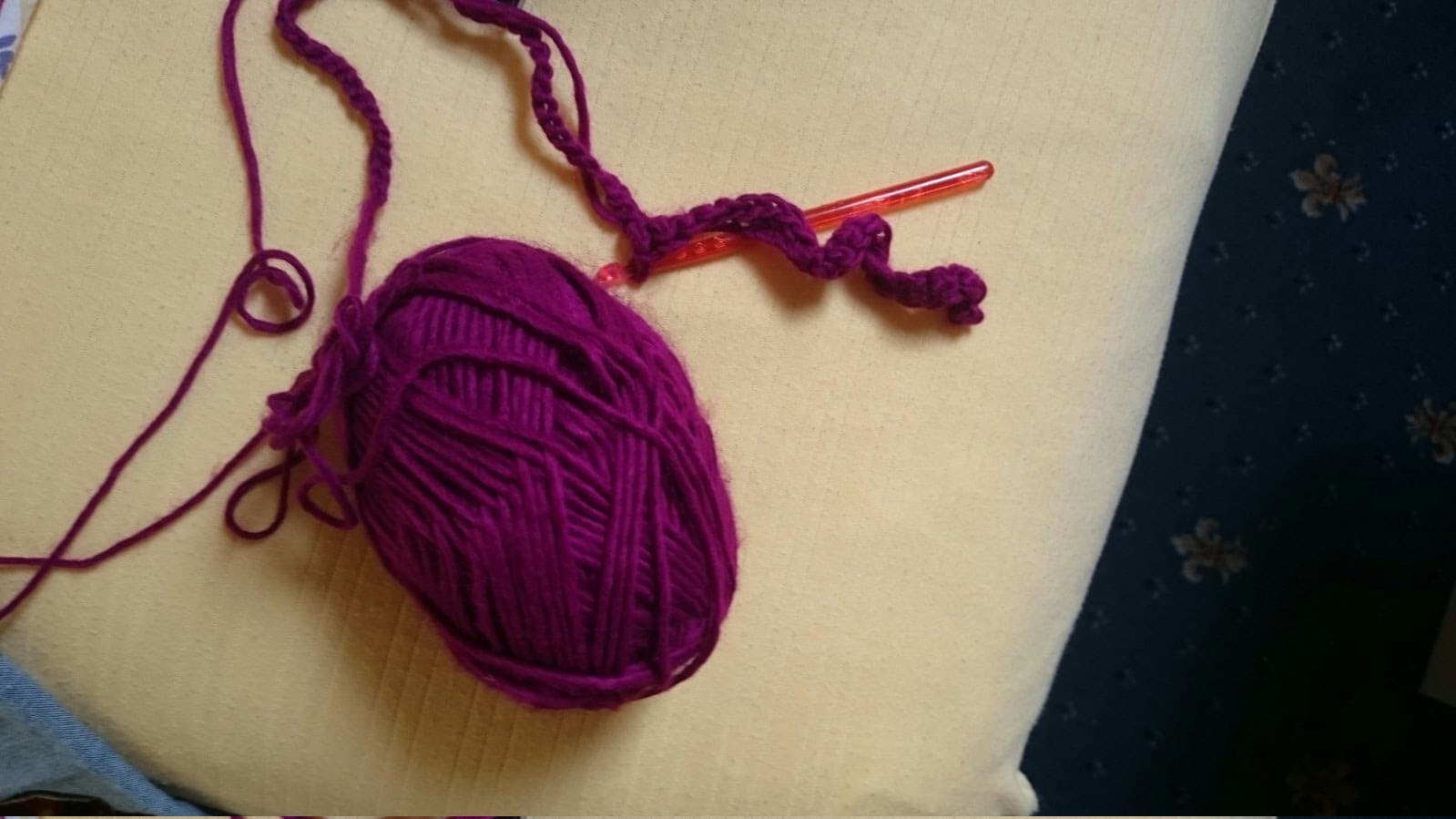}&
\includegraphics[width=0.2\textwidth]{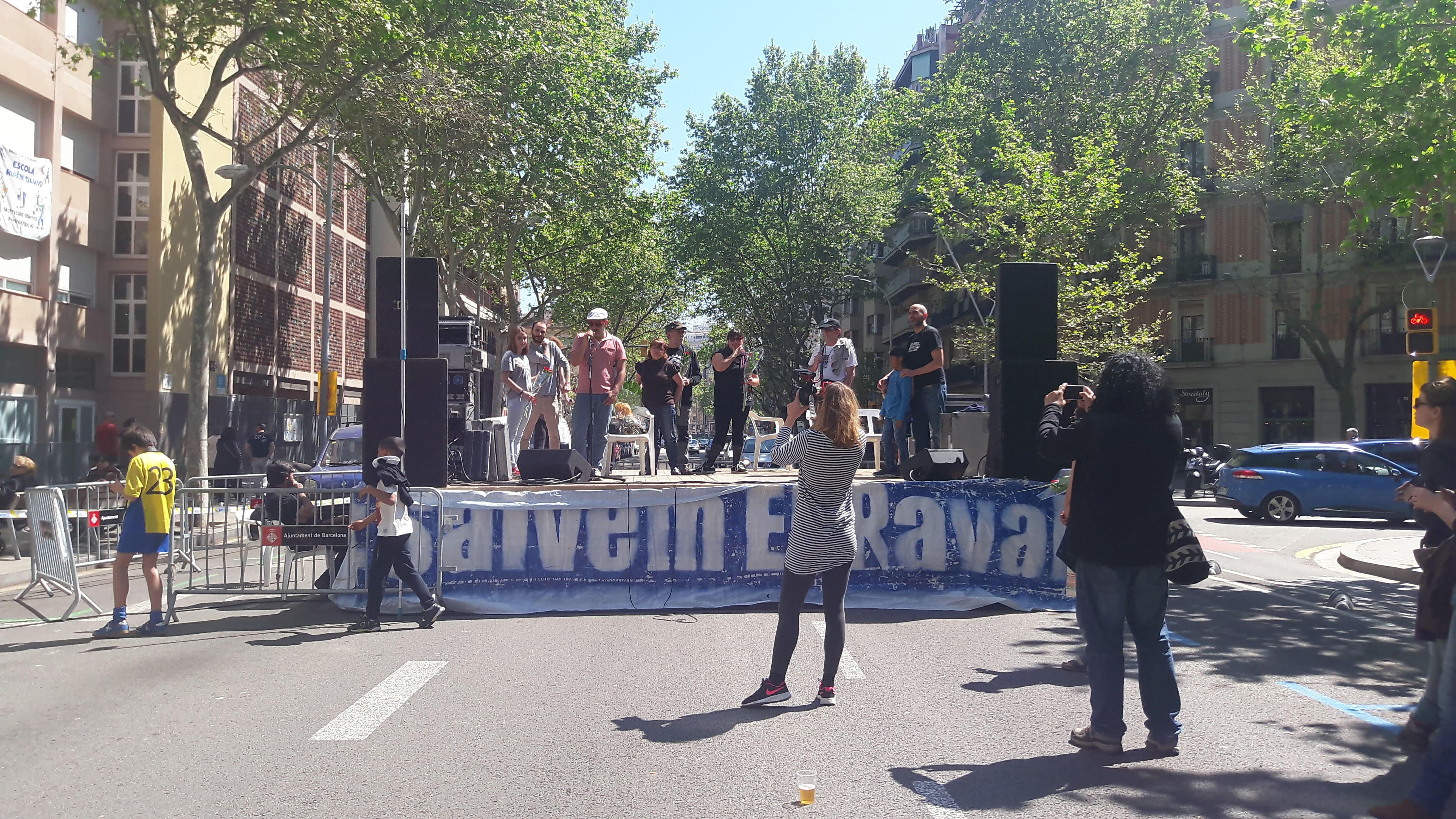}&
\includegraphics[width=0.2\textwidth]{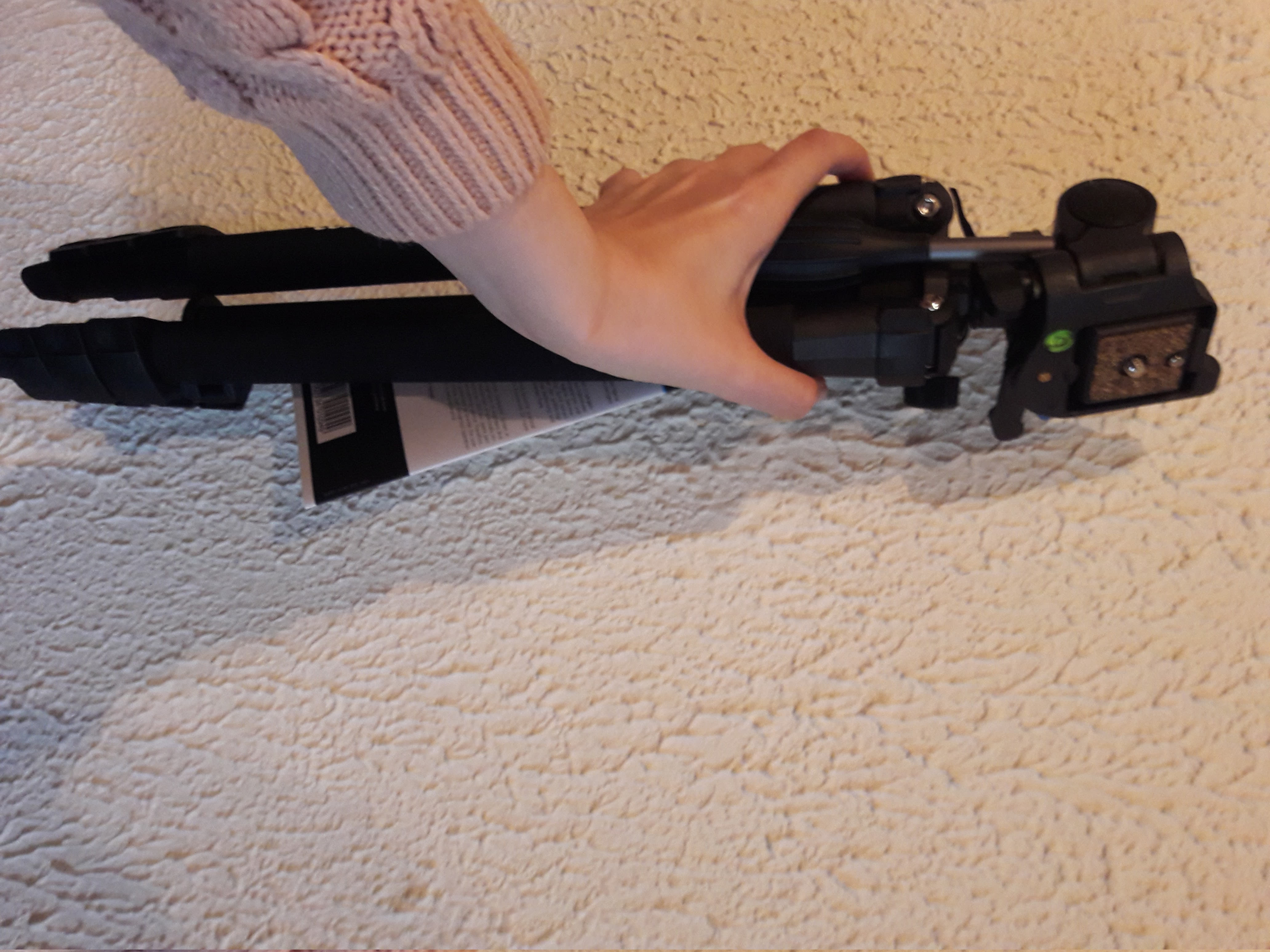}&
\includegraphics[width=0.2\textwidth]{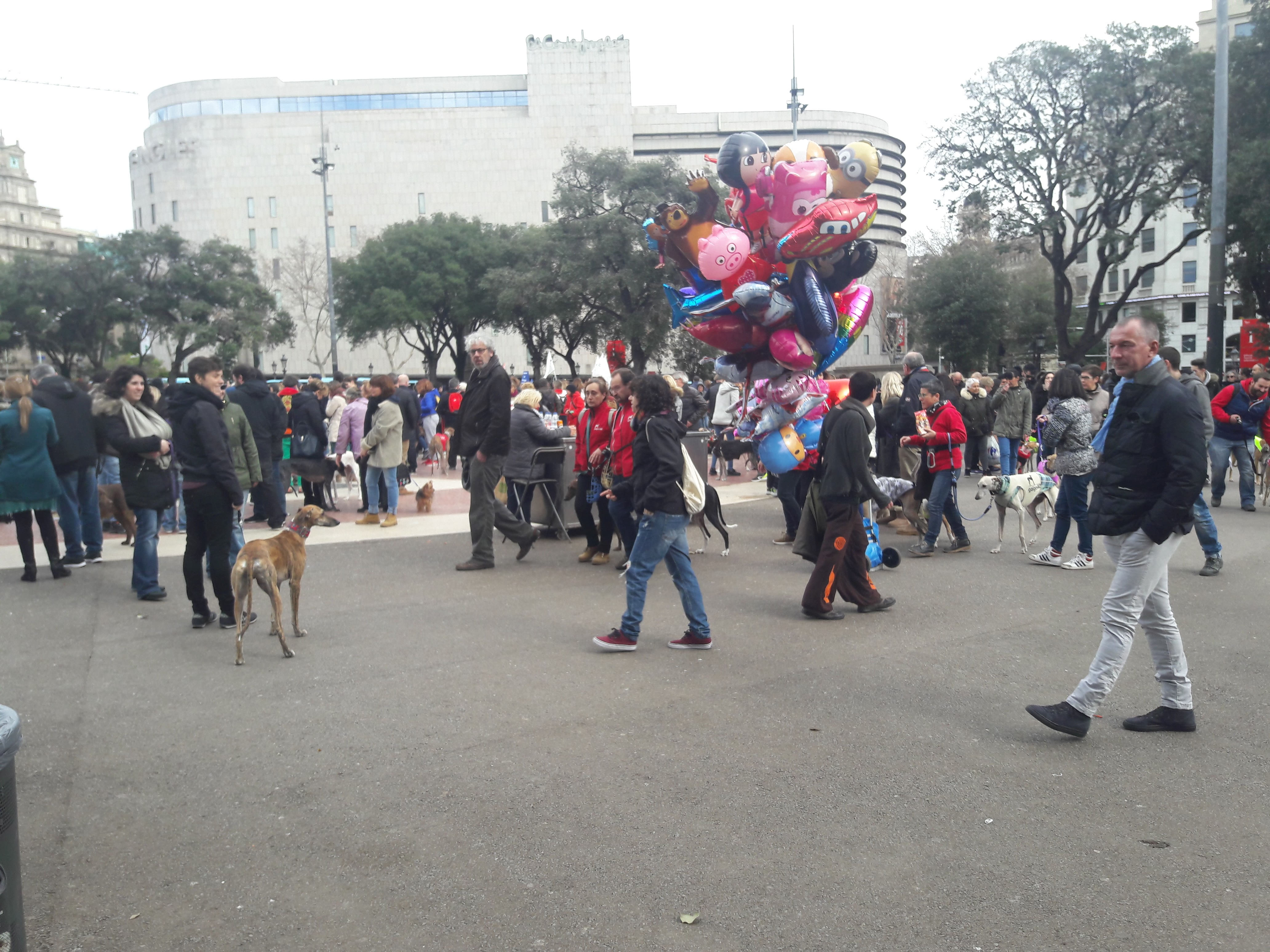}&
\includegraphics[width=0.2\textwidth]{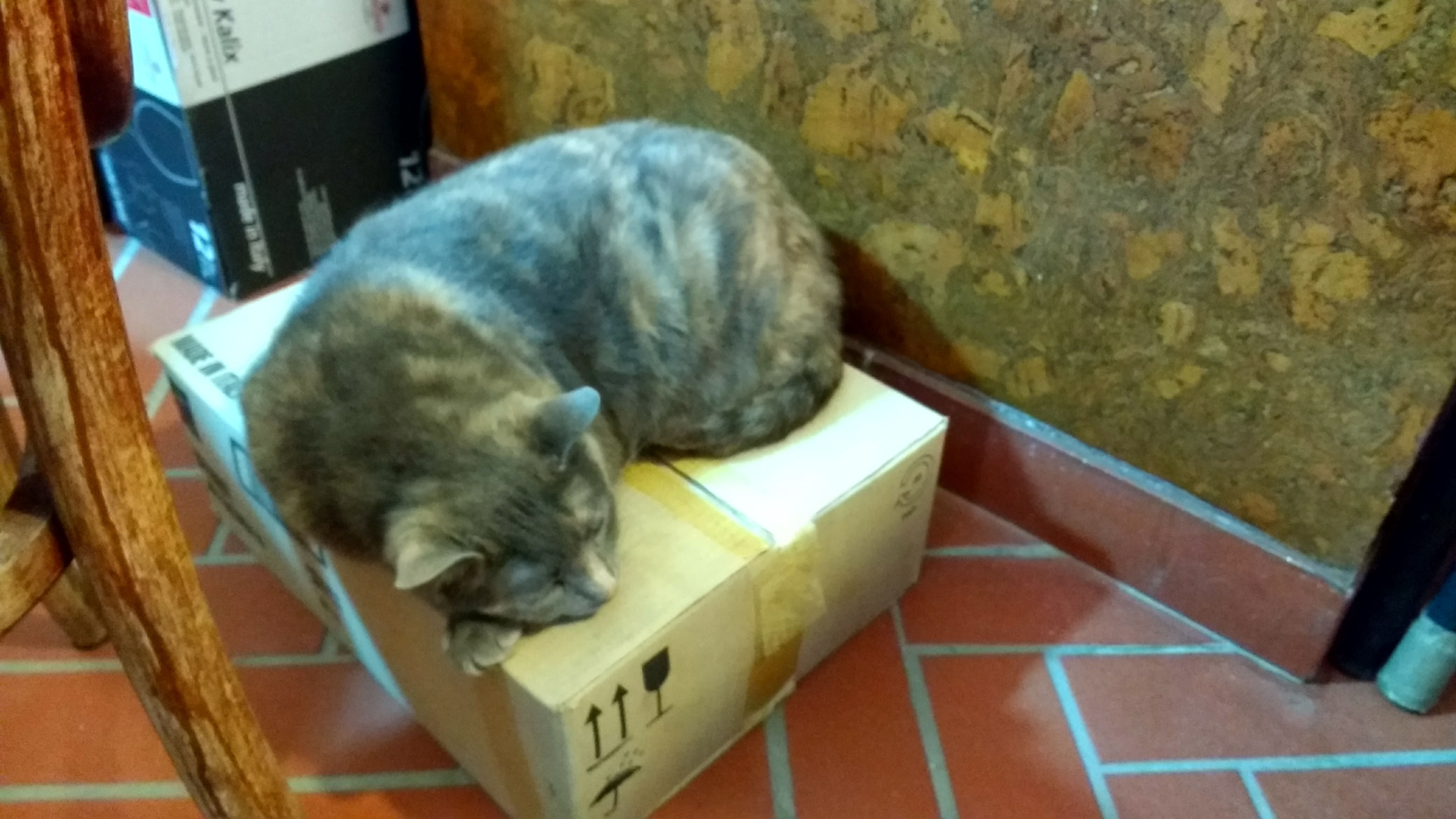}\\
archery & badminton & desert sand & skiing & painting
\end{tabular}
}
\caption{Example of images incorrectly classified: correct topic, but incorrect category (top), incorrect topic and incorrect category (bottom)}
\label{fig:negative}
\end{figure*}





\begin{table*}[!h]
\caption{Datasets and event classes used by state of the art algorithms}
\centering
\resizebox{1.0\textwidth}{!}{%
\begin{tabular}{ | c | c |c |c |c |c |c |c |c |c | }
\hline  \textbf{Holiday 1} &  \small{\cite{gu2013personal}} &  \textbf{Holiday 2}& \textbf{SocEID }&  & \multicolumn{2}{|c|}{\textbf{UIUC Sports}}  \\
 \small{\cite{bacha2016event,guo2015event}} &  \small{\cite{tsai2011compositional}} &  \small{\cite{cao2009image,cao2008annotating}}&  \small{\cite{tsai2011album}} &  \small{\cite{tang2011event}} & \multicolumn{2}{|c|}{ \small{\cite{ahsan2017complex}}}   \\ \hline
 Mardi gras & Beach fun &  Christmas & Birthdays  & Christmas & \multicolumn{2}{|c|}{Rowing  }\\ \hline  Thanksgiving & Graduation & Halloween & Graduations &  Halloween & \multicolumn{2}{|c|}{Badminton}\\ \hline   Christmas & Urban tour & Easter & Marathons/Races &   Valentines & \multicolumn{2}{|c|}{Polo}\\ \hline  Memorial day & Yardprk & Thanksgiving & Weddings & 4 July & \multicolumn{2}{|c|}{Bocce}\\ \hline New Years's Eve & Ball games & Independence's day & Protests  & Outdoor Sports  & \multicolumn{2}{|c|}{Snowboarding}\\ \hline Easter & Birthday  & New Years's Eve & Parades &  Birthday & \multicolumn{2}{|c|}{Croquet}\\ \hline  Valentine's day & Christmas  & Mardi gras & Soccer's matches & Beach & \multicolumn{2}{|c|}{Sailing}\\ \hline  Independence's day & Family time & Memorial day & Concerts &  Null event & \multicolumn{2}{|c|}{Rock climbing}\\ \hline  Halloween & Eating & San Patrick's day & &   & \multicolumn{2}{|c|}{Baseball}\\ \hline  San Patrick's day &  Skiing & Valentine's day&  &  & \multicolumn{2}{|c|}{} \\ \hline   & Wedding &  Labor day & &   & \multicolumn{2}{|c|}{}\\ \hline  & Null event & Mother's day &  &    &  \multicolumn{2}{|c|}{}\\  \hline  \textbf{PEC}   & \multicolumn{2}{|c|}{ \textbf{Rare Event Dataset}} & \multicolumn{4}{|c|}{\textbf{WIDER}  }         \\ 
 \small{\cite{bossard2013event,bacha2016event}} & \multicolumn{2}{|c|}{ \small{\cite{ahsan2017complex}}} & \multicolumn{4}{|c|}{ \small{\cite{ahsan2017complex}  }}      \\
\hline  Birthday &  J. Trudeau elected & Engagement parties  &  Parade & Soldier drilling & Photographers &  Traffic   \\ \hline   Children birthday & Election Trump & Boston red sox wins  & Handshaking & Spa & Raid &  Stock Market \\ \hline  Christmas &   Hurricane Katrina &  Humanity washed ashore & Demonstration & Sports fan  & Rescue &  Award ceremony   \\ \hline Concert & Hurricane Sandy   & Hot air balloon & Riot & Students Schoolkids & sports-Coach-Trainer   &  Ceremony  \\ \hline   Boat cruise &  Nepal earthquake &  Israel-Palestine conflict & Dancing & Surgeons  & Voter & Concerts   \\ \hline  Easter & 2012 summer Olympics  & Mali attacks  & Car accident & Waiter- Waitress &  Angler &  Couple\\ \hline Exhibition & Obama wins elections & Paris attacks  &  Funeral & Worker-Laborer  & Hockey &  Family group  \\ \hline  Graduation &  Columbia space shuttle disaster  & Royal wedding & Cheering & Running & People driving car &  Festival \\ \hline  Halloween &  Arab spring  &  Yemen civil war & Election campaign & Street battle  &  Traffic &  Picnic \\ \hline San Patrick's day  &  9/11 attacks & Thanksgiving & Russian airlines crashes Sinai & Press conference &  Basketball &    Shoppers\\ \hline         Road trip & Boston bombing & US invasion Afghanistan & People marching & Football & Interview &   Soldiers firing \\ \hline  Hiking &  Russian airstrikes Siria  & Meeting & Soccer  & Group &  Celebration or party & Soldiers Petrol \\ \hline Skiing &  Baby showers   &  &  Tennis &  Meeting & Dresses & Jockey \\ \hline  Wedding & Drones attacks Yemen Pakistan  &  &  Ice skating  & Greeting & Parachutist paratrooper & Matador bullfighter\\ \hline  &   &   & Gymnastic & Balloonist &  Aerobics & Row boat\\ \hline  & &  & Swimming & Car racing & &  \\  \hline \noalign{\vskip 0.25cm}   
\end{tabular}
}
\label{tab:SoAlabels}
\end{table*}

 Fig. \ref{fig:pwz} plots the topic-specific word distributions $P(w|z)$ computed on the training set and shows how different words have different probabilities of appearing in each topic. It can be observed that the three topic measures show consistent results for pLSA. The topic \textit{Sport and Exercise} is the less coherent whereas \textit{Nature and Panorama} and \textit{Art and visual} are the most coherent.
For comparison purpose, in Table \ref{lab:label}, we report the values of the topic coherence measures described in section \ref{subsub_validation}, for two other widely used topic models, namely  Latent Dirichlet allocation (LDA) \cite{blei2003latent} and Latent Semantic Analysis (LSA) \cite{dumais2004latent}. These results were obtained by using 
the Movie corpus, a Wikipedia subset, as external corpus in the \textit{gensim} Python library. As it can be observed, the average topic coherence in terms of UCI-score and Umass-score is much higher for pLSA than LSA and LDA, whereas the NPMI is slightly better for LSA. However, only three topics out of eight have a higher NPMI values, while most of them have very low values.
Furthermore, while with the results of pLSA each word distribution was automatically assigned to a different label, with the results of LDA and LSA two different word distributions were assigned to a same label. In particular, with LDA \textit{Art and Visual} and  \textit{Food and Beverage} were assigned twice to a same word distribution and the labels  \textit{Sport and Exercise} and  \textit{Interior and Decoration} were not assigned to any word distributions. Similarly, with LSA, the labels  \textit{Interior and Decoration} and  \textit{City and Architecture} were assigned twice to a same word distribution, whereas \textit{Party and People} and  \textit{Nature and Panorama} were not assigned to any word distribution.

\subsubsection{User study results}
We recruited thirty persons for the user study who were not involved with the data collection, and six of them were computer illiterate. In average, each photo collection has been evaluated by three different participants. The evaluations were slightly harsher depending on the participant background. We observed that people familiar with technology gave more feedback. 
Each participant evaluated at most three photos collections.
First, we asked people to draw down the categories into which they would like to organize their pictures. The most popular categories were: \textit{Friends}, \textit{Architecture}, \textit{Travel}, \textit{Panorama}, \textit{Selfies},  \textit{Food}, \textit{Documents}, \textit{Dogs}, \textit{Sport} (described with the favorite one such as Skatering). Less common categories were often related to the user job or to a particular hobby.

We compared our photo organization to the two most popular and automatic photo categorization systems, namely Eden and Google Photos. We evaluated two important aspects: a) categories organization, that is hierarchical organization versus just one layer classification, and b) image assignment to the categories. Regarding a), note that Eden has only 14 generic categories, whereas Google has 1100 subcategories. Our system has 8 generic topics and a total of 1311 subcategories (see Table \ref{tab:appComparison}).

In Table \ref{tab:user_study1}, we  report  the  results  of  the  user study,  together  with  the  corresponding  statistical  descriptors. Specifically, we applied a One Sample T-Test, whose null hypothesis was that the proposed system is ”better” or ”much better” than the control system in terms of organization or accuracy.  To the rates from “much worst” to “much better” we assigned scores from 1 to 5. An up-value larger or equal than 0.04 indicates that null hypothesis is true. Therefore, in terms of organization, our system is considered better that the two control systems, both by the photo owners and external evaluators in a statistical relevant way.  We also observed  that photo owners gave score slightly higher than external evaluators. 
In terms of accuracy, the null hypothesis is rejected by both the external evaluators and the photo owners when comparing to the accuracy of Google. However, the value of the mean is very close to the similarity value, that is 3, in both cases (2.77 and 2.84). With Eden Photos, we observed a different trend: the photo owners judged our system having slightly better accuracy than Eden and this result is statistically relevant, whereas external participants considered the accuracy of Eden slightly better (mean value 2.93)  but the null hypothesis cannot be rejected since the up pvalue is larger than 0.04. 
However, when analysing these results it must be taken into account that Google Photo classified in average 53.66$\%$ of the pictures, whereas the Eden Photos app 61.29$\%$ and our system 81.6$\%$. Indeed, we used an accuracy threshold only for the classification into topics so that all pictures fed to neural networks were considered in the user study without taking into account the classification confidence.

To evaluate the reliability of user study results, we computed the Intraclass Correlation Coefficient (ICC) one-way random commonly known as ICC1 (\cite{shrout1979intraclass}), since the raters who rate one user were not necessarily the same as those who rate another user. This design corresponds to a One-way Analysis of Variance (ANOVA) in which User is a random effect, and Rater is viewed as measurement error.  The ICC1 is a measure of absolute agreement and is sensitive to difference in means between raters. It is defined as follows: 
\begin{equation}
ICC1 = \frac{BMS + WMS}{BMS
(k - 1)WMS'},
\end{equation} where  $k$ is the number of judges rating each
target, (BMS) is a between-targets mean square and  (WMS) a within-target mean square.
As shown in Table \ref{tab:user_study1}, we obtained values above 0.4 when comparing both the organization and the accuracy to those of control systems, which is considered a good value for ICC1 (\cite{Shieh2016}), since ICC1 values are always lower that other intraclass coefficient measures not applicable to this context. These results were obtained by using the ICC package in R \footnote{https://cran.r-project.org/web/packages/ICC/index.html}.

It is very important to remark that Google Photos always classifies the images into a relatively small set of categories, in average 9 over the 40 users, although it is supposed to account for 1100 categories.
Furthermore, several participants observed that many categories such as \textit{sky},  \textit{flowers} or \textit{car} include all  pictures where even a small portion of sky (or a car or a flower in the background) is visible and therefore were judged ambiguous. Several other groups of categories such as \textit{food}, \textit{cooking}, \textit{recipes} and \textit{baking} were judged redundant. The same occurs for \textit{skyscrapers}, \textit{skylines}, \textit{towers}. Furthermore, the only category related to people that was found in the full testing set was \textit{selfies}. A number of participants commented that the categories of our proposed system better reflect the way they would organize their own pictures. However, some participant commented that it would be useful for our system to have intermediate categories. For instance between \textit{Animal and Pets} and  \textit{irish terrier}, it would be useful to have the intermediate category \textit{dog}. Although we did not show this in our user study, it is worth to observe that such intermediate classes are naturally provided by the synset associated to the subcategories. Additionally, other participants commented that it would be better to have the opportunity to choice for which topics to have subcategories and for which not. Others commented that sometimes the number of categories is too large given the number of pictures under the topic. These remarks suggest that the number of subcategories could be determined depending on the number of pictures under the main topic, that often unveils what the user like to capture with his/her smartphone. Finally, some participants commented that having information about the place is very important. We stress that many of the user suggestions could be easily integrated into the proposed approach by relying on additional information such as gps coordinates and EXIF metadata.
Overall, beside validating the proposed approach, the information collected through the user study will be useful for future developments.


\subsubsection{Qualitative results}

Fig. \ref{fig:positive} and \ref{fig:negative}  show examples of pictures correctly and incorrectly classified by our system, respectively. In particular,  in Fig. \ref{fig:positive} it is possible to appreciate the level of detail that can be achieved by our system. On the first row of Fig. \ref{fig:negative}, are shown examples of pictures that have been assigned to the right topic, but to the incorrect category, whereas on the second row are shown examples of pictures that have been assigned to a wrong topic. Since both topics \textit{People and Portraits} and \textit{Sports and Adventure} involve people, pictures with crow are easily wrongly assigned to \textit{Sports and Adventure}.

\subsubsection{Discussion}
Existing algorithms for event recognition from personal photo collections have focused on the detection of a limited set of social events (see Table \ref{tab:SoAlabels}). Even if the PEC dataset becomes a standard in the community, several other in-house datasets with very similar categories (see top part of Table \ref{tab:SoAlabels}) have been used in the literature (\cite{cao2008annotating,cao2009image,tsai2011album,tang2011event,tsai2011album,gu2013personal,ahsan2017complex}). However, in smartphone photo collections there are very few images that have been captured during the same event during a short period of time. Additionally, images captured by a smartphone have a large variability in terms of topics, so that those belonging to the category \textit{Parties and People} are just a (small) portion of them. For these reasons, a comparison with such methods would be unfair.

Although our work is closely related to image categorization for easing image access (seek, organize and understand images), it could serve also as basis for image retrieval. Indeed, since a multilabel approach is typically best suited for retrieval, the same image tags that we used as input to our system could be used together with the hierarchical labels to improve retrieval performance.
In addition,  time and localization metadata are easily available on a smartphone through GPS and Google Calendar. As demonstrated by the literature on event recognition, the use of this information would be extremely useful in detecting special events. For instance, this would help to retrieve and recognize rare events such the ones of the Rare Event Dataset  (see Table \ref{tab:SoAlabels}), that are currently not handled by the proposed system.  We leave this for future work.

\section{Conclusions} 
\label{conclusion}


This paper addressed the problem of organizing smartphone pictures into a set of topics and topic-related categories. The proposed approach first classifies images into eight topics by using an unsupervised generative approach that allows to account for their huge intra-class variability. Next, pictures are classified into a large number of categories by using a CNN approach.

User studies demonstrated that users prefer our two-levels classification with respect to a one-level classification provided by widely used photo organization systems such as Eden Photos and Google Photos.  
The proposed approach could be easily integrated in a retrieval system that relies on both semantic tags, time and location metadata to retrieve all images corresponding to the user query.  With the goal of encouraging further research on smartphone picture organization, we make available a dataset of smartphone pictures from 40 persons.

\section*{Acknowledgments}

We kindly acknowledge the European Network on Integrating Vision and Language (iV$\&$L Net) ICT COST Action IC1307 for the support of a Short Term Visit of one of the authors to Imagga Ltd. The authors are very grateful to Stavri Nikolov for inspiring conversations. We also gratefully acknowledge the support of NVIDIA Corporation with the donation of the GPUs used in our research.
Finally, we thank two anonymous reviewers for their helpful and constructive comments that greatly contributed to improving the final version of the paper.

This work was partially funded by TIN2015-66951-C2-1R, SGR-1219, Grant 20141510 (Marat\'{o} TV3),  CERCA Programme / Generalitat de Catalunya and \textit{ICREA Academia} grant.

\bibliographystyle{unsrt}  

\bibliography{references}

\end{document}